\documentclass[lettersize,journal]{IEEEtran}
\usepackage{amsmath,amsfonts}
\usepackage{tikz}
\usetikzlibrary{arrows, positioning}
\usepackage{algorithmic}
\usepackage{algorithm}
\usepackage{array}
\usepackage[caption=false]{subfig}
\usepackage{textcomp}
\usepackage{stfloats}
\usepackage{url}
\usepackage{verbatim}
\usepackage{graphicx}
\usepackage{cite}
\usepackage{amsmath,amsfonts,bm}

\def\Figr#1{Fig.~\ref{#1}}

\def\Eqr#1{Eq.~\eqref{#1}}

\def\Algref#1{Algorithm~\ref{#1}}

\def\1{\bm{1}}

\def\rz{{\textnormal{z}}}

\def\rvc{{\mathbf{c}}}

\def\rvr{{\mathbf{r}}}

\def\rvw{{\mathbf{w}}}
\def\rvx{{\mathbf{x}}}

\def\rvz{{\mathbf{z}}}

\def\ervc{{\textnormal{c}}}

\def\ervr{{\textnormal{r}}}

\def\ervw{{\textnormal{w}}}
\def\ervx{{\textnormal{x}}}

\def\ervz{{\textnormal{z}}}

\DeclareMathAlphabet{\mathsfit}{\encodingdefault}{\sfdefault}{m}{sl}
\SetMathAlphabet{\mathsfit}{bold}{\encodingdefault}{\sfdefault}{bx}{n}

\def\sC{{\mathbb{C}}}

\def\sH{{\mathbb{H}}}

\def\sR{{\mathbb{R}}}
\def\sS{{\mathbb{S}}}

\def\sX{{\mathbb{X}}}
\def\sY{{\mathbb{Y}}}
\def\sZ{{\mathbb{Z}}}

\begin{document}

\title{A Multi-stage Constrained Optimization Framework for Data-driven Problems}

\author{Ye~Shi,~\IEEEmembership{Student~Member,~IEEE}
\thanks{Ye Shi~(shi349@purdue.edu) is with the Elmore Family School of Electrical and Computer Engineering, Purdue University, West Lafayette, IN 47907, USA.}
\thanks{Manuscript received TBD; revised TBD.}}


\maketitle

\begin{abstract}
Variational Autoencoders (VAEs) have emerged as a powerful tool for transforming high-dimensional, often noisy data, into a compact latent representation, thereby enabling more tractable optimization.
However, three key challenges persist in VAE-based constrained optimization: (i) ensuring effective sampling within the latent space, (ii) identifying active decision variables that truly impact the objective and constraints, and (iii) enforcing hard constraints without destabilizing training. 
In this paper, we propose a novel Multi-stage Constrained Optimization Framework (MCOF) to overcome these challenges. 
First, we introduce an entropy-constrained VAE (EC-VAE) coupled with a feature selector to systematically embed critical objective and constraint information into a subset of latent variables. 
Next, a Uniform Transformation (UT) module refines the latent representation by converting it into a uniform distribution, thus mitigating issues of posterior collapse and Gaussian mixture biases. 
To solve the reformulated constrained optimization problem, we develop a constraint-priority filter method (CPFM), which treats constraint satisfaction as a high-priority objective and demonstrably improves feasibility relative to standard penalty or Lagrangian-based approaches. 
We validate MCOF on a synthetic optimization problem and a drug discovery application. Results show that MCOF reliably converges to feasible, near-optimal solutions while maintaining computational efficiency, highlighting its potential as a robust and general-purpose methodology for data-driven constrained optimization.
\end{abstract}


\section{Introduction}
\IEEEPARstart{V}{ariational} autoencoders (VAEs)~\cite{kingmaIntroductionVariationalAutoencoders2019} are a powerful tool for reconstructing data representations in latent spaces, enabling the modeling of both discrete and continuous distributions~\cite{saseendranShapeYourSpace2021}.
Researchers can apply various optimizers to optimization problems transformed by VAEs, including quantum algorithms~\cite{wilsonMachineLearningFramework2021}, evolutionary algorithms~\cite{bentleyCOILConstrainedOptimization2022}, Bayesian optimization~\cite{griffithsConstrainedBayesianOptimization2019}, etc.
In addition to common continuous optimization,
the transformation of discrete spaces into continuous latent spaces is particularly superior in complex combinatorial optimization scenarios~\cite{hottungLearningHeuristicsCombinatorial2023}, such as those encountered in molecular discovery and drug design~\cite{notinImprovingBlackboxOptimization2021, maConstrainedGenerationSemantically2018, liuConstrainedGraphVariational2018,griffithsConstrainedBayesianOptimization2019, gomez-bombarelliAutomaticChemicalDesign2018}.
This is accomplished by embedding original categorical data samples into graphs~\cite{liuConstrainedGraphVariational2018} or texts~\cite{gomez-bombarelliAutomaticChemicalDesign2018}, subsequently transformed by VAEs into continuous latent spaces.
By simply appending the decoder of VAEs to the original objective function, optimizers can then sample from these continuous latent spaces, thereby overcoming the limitation inherent in discrete search spaces.
Moreover, certain studies~\cite{wilsonMachineLearningFramework2021} have explored the construction of discrete latent spaces tailored to the needs of specialized optimizers.
By reformulating the data representation, VAEs exhibit significant potential for optimizing a range of complex problems.
However, several challenges persist. 
In data-driven black-box optimizations, objectives and constraints are often implicit, and identifying which decision variables actually drive the performance remains nontrivial. 
Furthermore, even once relevant variables are found, determining how to efficiently navigate latent spaces without violating hard constraints requires careful algorithmic design. 
These considerations motivate the three interrelated challenges examined in this paper.

\textit{Challenge 1: Sampling in Latent Space.}
A primary difficulty lies in effectively sampling from the VAE's latent space in a way that respects feasibility requirements. 
Although VAEs can learn powerful representations from raw data, they do not inherently account for objective functions or constraints, which can lead to generated solutions that are either suboptimal or infeasible~\cite{kotaryEndtoEndConstrainedOptimization2021a, daiDiagnosingEnhancingVAE2019b}. 
Nevertheless, the inherent ability of VAEs to serve as surrogate models remains advantageous, as they can rapidly adapt to ill-defined or multi-objective problems by learning latent structures directly from data~\cite{zhuBayesianDeepConvolutional2018, parekhVariationalAutoencoderBasedMetamodeling2022, liNeuralArchitectureOptimization2020}. 
For well-studied problems, pre- or post-hoc corrections can refine solutions effectively~\cite{liuConstrainedGraphVariational2018,bentleyCOILConstrainedOptimization2022}. 
However, in underexplored problems---especially when training data is imbalanced or biased---the latent space may exhibit irregular distributions, compromising the convexity of the search space for surrogate models~\cite{notinImprovingBlackboxOptimization2021}.
To balance generative flexibility with the need for validity, hybrid approaches often integrate domain-specific rules during decoding or apply post-hoc optimizers for correction, albeit at the cost of increased complexity and tuning.

\textit{Challenge 2: Determining Active Decision Variables.}
A second challenge emerges in data-driven black-box settings, where objectives and constraints are implicit, and not all variables necessarily affect the outcome. 
Identifying \emph{active} decision variables---those that genuinely influence performance or feasibility---is critical before building any surrogate model. 
The type of constraints involved (linear or non-linear, equality or inequality) significantly affects the complexity of enforcing them. Early approaches, such as DC3~\cite{dontiDC3LearningMethod2021}, rigorously classify decision variables based on constraint type and use gradient descent to ensure feasibility. 
More recent research leverages Lagrangian duality, embedding constraints into the loss function via multipliers~\cite{fiorettoLagrangianDualityConstrained2021}, or employs dual networks that adversarially guide primal networks using constraint violations~\cite{parkSelfSupervisedPrimalDualLearning2023}. 
Despite such advancements, systematic methods for identifying and enforcing active variables within VAE-based optimization remain underexplored.

\textit{Challenge 3: Constrained Optimization Algorithms (Especially Cooperating with Hard Constraints).}
Finally, even after pinpointing active decision variables and learning suitable latent representations, selecting or designing an optimization algorithm that can accommodate hard constraints poses its own difficulties. 
Traditional methods are often classified as gradient-based or gradient-free methods. 
Gradient-based techniques exploit function derivatives and treat constraints as penalties, a practice which can lead to suboptimal solutions in non-convex, over-parameterized deep models~\cite{liuLossLandscapesOptimization2022, choongOptimizingVariationalGraph2020}. 
Conversely, gradient-free methods rely on sampling and can be inefficient in high-dimensional or complex spaces. 
Recent works propose various strategies for constrained optimization in deep-learning contexts: DC3~\cite{dontiDC3LearningMethod2021} ensures feasibility through iterative gradient updates, while Lagrangian frameworks integrate constraints into training objectives via multipliers~\cite{fiorettoLagrangianDualityConstrained2021}. 
Yet, VAEs require special attention when dealing with hard constraints due to their generative flexibility. 
Approaches to reconciling strict feasibility with VAE-based sampling include domain-specific decoder regularization and post-hoc correction~\cite{kotaryEndtoEndConstrainedOptimization2021a}, underscoring the complexity of finding an efficient balance between strict constraint satisfaction and the exploratory potential of VAEs.

In summary, sampling in latent spaces, determining active decision variables, and cooperating with hard constraints remain pivotal and interlinked challenges in VAE-based constrained optimization. 
Progress in these areas promises to broaden the applicability of VAEs to an ever-increasing range of real-world optimization problems, particularly those defined by implicit or dynamically evolving constraints.

In this paper, we propose a novel \textit{multi-stage constrained optimization framework} (MCOF) to address key challenges in data-driven black-box optimization. 
Our framework comprises two main phases: a training phase and an optimizing phase. 
During the training phase, we construct a latent-based surrogate model of the original problem in two stages. 
In Stage 1 (Targets Embedding), we tackle the challenge of identifying active decision variables by introducing an \textit{entropy-constrained VAE} (EC-VAE).
The EC-VAE imposes a lower bound on the entropy of latent variables, coupled with a feature selector that extracts the most informative latent dimensions.
In Stage 2 (Refining the Decoder and Regressor), we address the challenge of sampling in latent space by incorporating a \textit{Uniform Transformation (UT) module} that converts irregular latent distributions into a uniform one.
Thereby, Stage 2 enhances downstream performance and recovers convexity in the latent space.
Once this surrogate model is established, the optimizing phase generates solutions. 
Stage 3 employs a \textit{constraint-priority filter method} (CPFM) that recasts constrained optimization as a multi-objective problem finding the Pareto frontier between the primary objective and a violation function. 
Stage 4 completes the latent solution by sampling the remaining unselected dimensions.
This enables the generation of diverse solutions within the original data space while maintaining constraint satisfactions and optimality.

Our proposed MCOF framework contributes significantly to the field of constrained optimization by seamlessly integrating VAE-based generative modeling with robust constraint handling. 
It balances generative flexibility with constraint validity, enabling efficient and feasible solution generation in high-dimensional, data-driven, black-box problems.
Our approach is particularly impactful for under-explored, ill-defined, or multi-objective problems where conventional optimization techniques struggle to maintain validity.
By validating MCOF on both synthetic and real-world applications, including drug discovery, we demonstrate its potential to reliably converge to feasible, near-optimal solutions while maintaining computational efficiency. 
The versatility and general-purpose design of MCOF allow it to adapt to various data-driven optimization scenarios, making it particularly valuable in fields such as operations research, engineering design, and pharmaceutical research.
This work not only provides a systematic approach to overcoming the inherent challenges of VAE-based optimization but also establishes a flexible framework applicable to complex, real-world problems and benefits to research in disentangled representation and generative models.
Consequently, MCOF opens up new possibilities for leveraging VAEs in solving an expanding range of real-world optimization challenges.

The remainder of this paper is organized as follows. 
Section~\ref{sec:related} reviews related work on latent-space sampling, constraint imposition, and inverse optimization methods. 
Section~\ref{sec:approach} formulates the data-driven problem and introduces our proposed MCOF in detail. 
Section~\ref{sec:experiment} presents experimental results. 
Finally, Section~\ref{sec:conclusion} concludes and summarizes the paper.

\section{Related Works}\label{sec:related}
This section presents a comprehensive review of the literature addressing key challenges in sampling in VAEs, constraint optimization in deep learning and inverse problems and optimization within the domain of VAE-based optimization. 
These aspects are critical as they significantly influence the effectiveness and efficiency of VAE models in various applications, from generative tasks to complex optimization scenarios.

\subsection{Sampling in VAEs}
\begin{figure}
    \centering
    \includegraphics[width =0.9 \columnwidth]{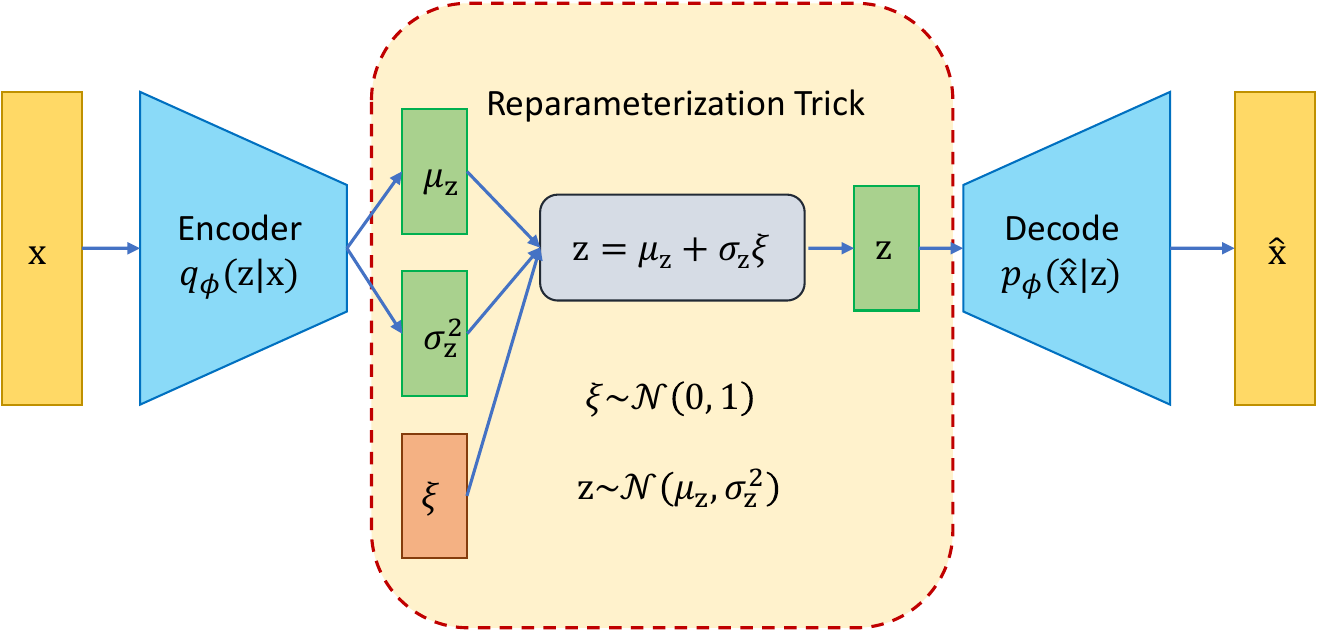}
    \caption{VAE Structure. The reparameterization trick separates the encoder output into mean $\mu_{\rz}$ and variance $\sigma_{\rz}^2$, which are then combined with an independent random variable $\xi$ to construct the latent variable $\rz$.}
    \label{fig:VAE}
\end{figure}

Sampling is currently considered a significant challenge in VAEs~\cite{chadebecPythaeUnifyingGenerative2023}.
It leads to poor quality in generated samples when randomly sampling the latent space.
This issue has been long recognized by researchers who have continually refined their understanding of it. 
Initially, it was thought that this problem stemmed from the uncertainty introduced by the reparameterization trick. 
This trick enables the back-propagation of learning mean $\mu_{\rvz}$ and variance $\sigma_{\rvz}$ of the latent variable $\rvz$ onto a normal distribution $p(\rvz)$, as illustrated in Fig.~\ref{fig:VAE}.
An external random variable $\xi \sim \mathcal{N}(0,1)$ facilitates this process, ensuring each latent variable value $\rvz$ follows a Gaussian distribution.
An intuitive response has been to make the latent representation more deterministic through transformations of the latent space,
explored in depth through flow-based approaches such as normalized flow~\cite{rezendeVariationalInferenceNormalizing2016},
autoregressive flow~\cite{kingmaImprovingVariationalInference2017},
and Householder flow~\cite{tomczakImprovingVariationalAutoEncoders2017}.

Subsequently, a phenomenon known as posterior collapse~\cite{burdaImportanceWeightedAutoencoders2016} was observed,
where the correlation analysis between input data and latent representation indicated that in a sufficiently dimensioned latent space,
some variables failed to encode any informative data features,
leading to a collapse of the posterior distribution of the encoder~\cite{wangPosteriorCollapseLatent2021}.
This explained the poor generative performance when randomly sampling in a high-dimensional latent space, 
the learned representation was over-parameterized and introduced excessive noise from $\xi$. 
Numerous studies have since been conducted to increase the number of active variables that encode informative data features. 
In addition to flow-based methods, another prominent approach involves regulating the latent variables, 
such as employing total correlation~\cite{kimDisentanglingFactorising2018,chenIsolatingSourcesDisentanglement2018}
and Gaussian mixture models~\cite{dilokthanakulDeepUnsupervisedClustering2017a,ghoshVariationalDeterministicAutoencoders2020, saseendranShapeYourSpace2021}.

More recently, researchers have identified the misalignment between the encoding posterior distribution and the decoding prior distribution~\cite{shenImprovingVariationalEncoderDecoders2018},
particularly when the learned features became sparse, categorical, or discrete.
The encoded latent distribution often forms an irregular distribution in the latent space,
whereas the decoder is typically designed only for strictly normal distributions~\cite{safarGenerativeModelsIrregular2021, naimanGenerativeModelingRegular2023}.
Several solutions have since been developed to correct this alignment, 
including pairing distributions between the encoder and decoder~\cite{koike-akinoAutoVAEMismatchedVariational2022},
identifying mismatches in representation~\cite{cemgilAutoencodingVariationalAutoencoder2020a}, 
and coupling the posterior and prior distributions~\cite{haoCoupledVariationalAutoencoder2023}.

For our demands in VAE-based optimization, 
invertibility and determinism of the latent representation are crucial for constructing downstream models of objectives and constraints,
as well as for identifying an optimal solution.
However, achieving these two properties simultaneously is challenging.
Drawing inspiration from recent advancements like the denoising diffusion probabilistic model~\cite{hoDenoisingDiffusionProbabilistic2020a} and the hierarchical VAE~\cite{vahdatNVAEDeepHierarchical2020} in computer vision, 
the VAE can be conceptualized as an invertible denoising solver.
To address this, our proposed UT module is implemented as the second stage modeling structure~\cite{daiDiagnosingEnhancingVAE2019b} in our training phase,
which facilitates an internal latent space for additional operations.

\subsection{Constrained Optimization in Deep Learning}
In deep learning research, constrained optimization often serves as a means to regulate model parameters or latent representations.
Especially in pursuing disentangled representations~\cite{chenIsolatingSourcesDisentanglement2018, kimDisentanglingFactorising2018}, latent-based soft constraints help suppress shared encoding features.
However, identifying the meaningful latent features still relies heavily on manual inspection by comparing the representation and the data features.
Therefore, disentangled representation has been facing a strong accusation as a result of inductive bias from the known facts~\cite{locatelloChallengingCommonAssumptions2019}.
In our framework, we also adopt an entropy constraint of latent variables as a regulation of latent representations to boost the latent-variable entropy.
We further introduce a feature selector, which eliminates inductive bias by automatically pinpointing the relevant latent variables. This offers a clearer, more rigorous form of disentangled representation.

When it comes to imposing constraints in deep-learning-based optimization, Lagrangian primal-dual approaches remain a popular choice~\cite{kotaryEndtoEndConstrainedOptimization2021a}. 
These methods incorporate constraints through multipliers, effectively penalizing violations.
Despite their appeal, finding stable multipliers in over-parameterized models poses a significant challenge in convergence---especially when constraints are implicit~\cite{marquez-neilaImposingHardConstraints2017}. 
Even a stable solution may still fail to strictly enforce constraints, reducing most primal-dual methods to soft-constrained outcomes~\cite{parkSelfSupervisedPrimalDualLearning2023,nandwaniPrimalDualFormulation2019}.

An alternative is to limit constraints to explicit, closed-form functions in decision variables.
Several works adopt a filtering mindset, discarding infeasible solutions post-generation. 
For instance, COIL \cite{bentleyCOILConstrainedOptimization2022} constructs a latent space from valid data only, 
while DC3 \cite{dontiDC3LearningMethod2021} persistently checks and corrects generated solutions if they violate constraints.
Both essentially treat constraints as filters that prune invalid inputs or outputs like many domain specified methods~\cite{griffithsConstrainedBayesianOptimization2019, maConstrainedGenerationSemantically2018, liuConstrainedGraphVariational2018}.
Drawing inspiration from this filtering paradigm, we reinterpret constrained optimization as a multi-objective process that searches the Pareto frontier of the primary objective and the satisfaction of constraints. 
This viewpoint aligns with classic filter methods \cite{fletcher_brief_2006}, originally designed for sequential quadratic programming.
Our proposed CPFM extends this idea by explicitly framing objectives and constraints as separate targets, then using a constraint priority mechanism to guarantee feasibility while still guiding solutions toward optimal performance. 
This strategy allows us to achieve genuine hard-constrained solutions, even in complex data-driven environments~\cite{pourmohamad_statistical_2020}.

\subsection{Inverse Problems and Optimization}
The inverse problems aim to estimate unknown parameters using observed data~\cite{kamyabDeepLearningMethods2022}.
VAEs have emerged as a powerful tool in addressing ill-posed inverse problems, where solutions may not exist, be unique, or remain stable. 
They have been applied in various contexts, such as image denoising~\cite{imDenoisingCriterionVariational2017},
diffusion processes~\cite{hoDenoisingDiffusionProbabilistic2020a},
and solving partial differential equations~\cite{gohSolvingBayesianInverse2021}. 
These studies commonly use VAEs to derive the posterior distribution of real data, from which solutions are sampled.

However, many of these applications do not fully leverage the intrinsic capability of VAEs as pairs of injective inverse functions.
Ideally, when a VAE perfectly reconstructs its input,
$\hat{\rvx} = \rvx$, 
it serves as two injective functions:
the encoder $\rvz = E(\rvx)$ and the decoder $\rvx = D(\rvz)$. 
Researchers~\cite{almaeenVariationalAutoencoderInverse2021} have introduced an inverse mapping between embedded labels in the latent space and real data. 
For example, recent developments in motor design~\cite{xuElectricMachineInverse2023} incorporate the design target within the latent space to address design problems involving non-unique solutions. 
Nevertheless, the direct exploitation of inverse relationships between latent variables and input data is still not widely explored.

Data-driven optimization problems in deep learning, often non-convex and over-parameterized, typically result in multiple invalid solutions~\cite{baiRecentAdvancesAdversarial2021}. 
This characterizes such issues as inverse optimization problems~\cite{chanInverseOptimizationTheory2023}:
\begin{equation}
    \begin{aligned}
        \min_x &\quad L\left(y_{obs}, \hat{f}(x)\right),\\
            \text{subject to} &\quad \hat{h}_i(x) = 0\quad \text{for all } i,\\
 &\quad \hat{g}_j(x) \leq 0 \quad \text{for all } j,
    \end{aligned}
\end{equation}
where $L(\cdot)$ denotes the loss function comparing the observed variable $y_{obs}$ with the estimation $\hat{f}(x)$,
$\hat{h}_i(x)$ and $\hat{g}_j(x)$ are the estimations of equality and inequality constraints.

Initial research on input optimization~\cite{szegedyIntriguingPropertiesNeural2014} showed that gradient-based methods often fail to find regular inputs that maximize the output, 
suggesting that such problems are ill-suited to traditional optimization methods. 
Consequently, global search and non-gradient algorithms have become popular for their efficacy in managing these challenges,
though they resemble solving a black-box problem and are time-consuming~\cite{notinImprovingBlackboxOptimization2021}. 
Recent work~\cite{gonzalezSolvingInverseProblems2022} has explored constructing a quadratic loss function $L(\rvx, \rvz)$ that exhibits weak bi-convexity, tackled through a posterior maximization process. 
Despite these advances, solving latent-based inverse optimization problems remains largely underexplored.

\section{Proposed Approach}\label{sec:approach}
\subsection{Problem Definition and Formulation}

Our proposed Multi-stage VAE-based Constrained Optimization Framework is designed to address black-box optimization problems where sufficient data are available.
Considering a given dataset as $\{\sX, \sY, \sC_E, \sC_I\}$ where $\sX$ represents the set of decision variable vectors $\rvx \in \mathbb{R}^d$, $\sY$ denotes the set of scalar objective values $y \in \mathbb{R}$, $\sC_E$ is the set of equality constraint vectors $\rvc_E \in \mathbb{R}^{m_E}$, and $\sC_I$ includes the inequality constraint vectors $\rvc_I \in \mathbb{R}^{m_I}$.  
Considering solving the constrained optimization problem as follows.
    \begin{equation}\label{eq:constrainedOptimization}
\begin{aligned}
    \min_{\rvx \in \mathbb{R}^d}   &\quad y, \\
    \text{subject to} &\quad\rvc_E = 0,\\
 &\quad \rvc_I \leq 0 .
\end{aligned}   
\end{equation}
Both the objective $y = f(\rvx)$ and the constraints---$\rvc_E = h(\rvx)$ for equality constraints and $\rvc_I = g(\rvx)$ for inequality constraints---are black-box functions, meaning their explicit forms are unknown and can only be estimated through data observations.
The decision variables $\rvx$ may be high-dimensional and noisy, and some components might be irrelevant to the optimization.

To address these challenges, we propose the MCOF framework consisting of two phases---training phase and optimizing phase.
Training phase aims to build an optimizable latent-based surrogate model and a latent solution decoder for interpreting the solution to the original decision space.
Optimizing phase aims to solve the latent-represented problem and generate solutions from the optimal solutions.

\begin{figure}

\centering
\includegraphics[scale = 0.33]{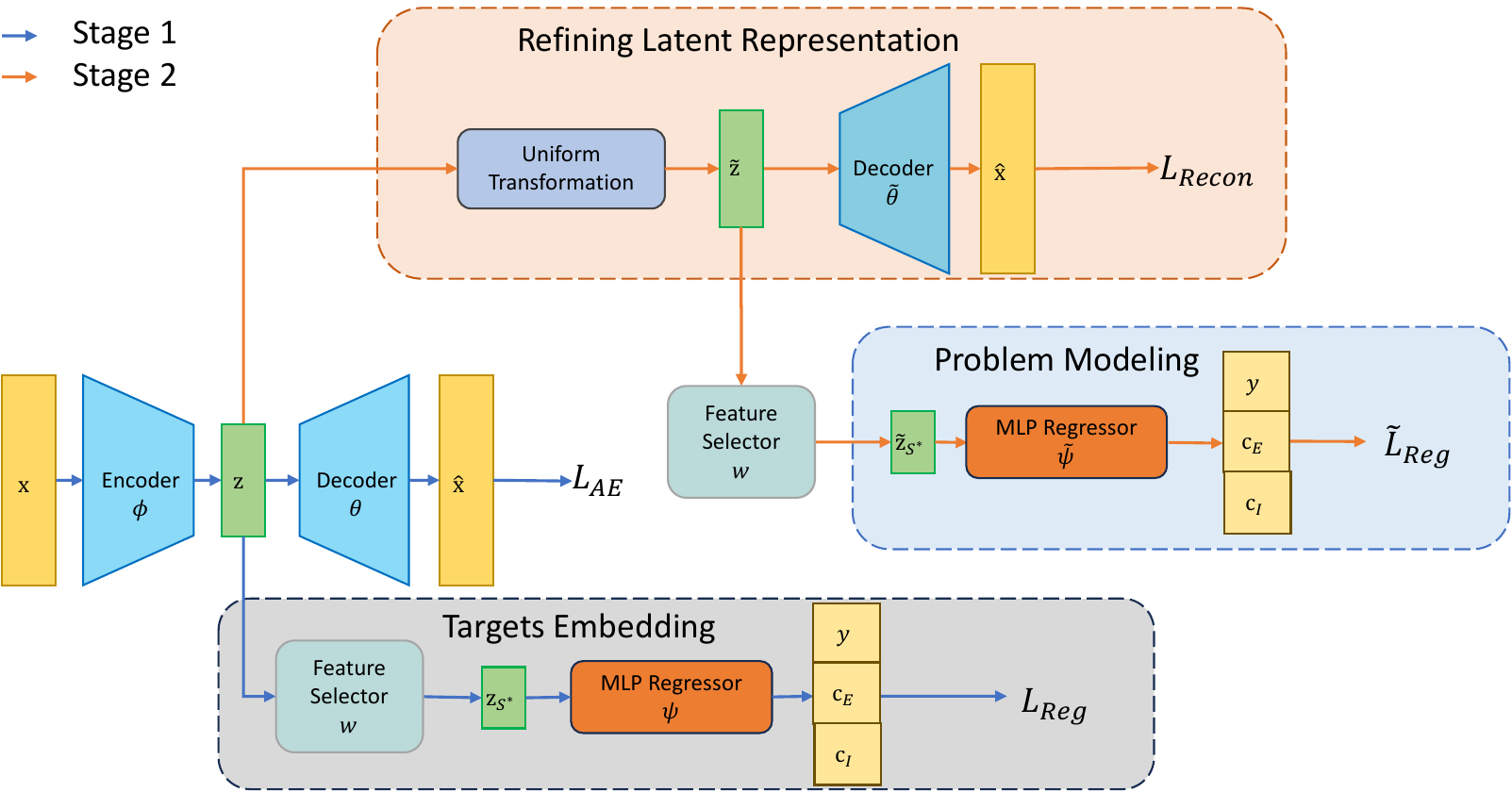}
    \caption{Diagram of Training Phase. 
The color-coded pipelines illustrate the processes within each stage. 
In Stage 1, the target features are embedded into the latent space using an entropy-constrained VAE with a feature selector. 
In Stage 2, the latent representation is refined, and a problem model is built using the selected latent features. 
}\label{fig:training}
\end{figure}

In the \textit{training phase} as Fig.~\ref{fig:training}, we transform the problem from the original decision space $\sX$ to a latent space $\sZ$ using an entropy-constrained VAE (EC-VAE).
In Stage 1, we embed the target features---the objective value $y$, equality constraints $\rvc_E$, and inequality constraints $\rvc_I$---into a selected subset of latent features $\rvz_{\sS^*}$ of the VAE.
The subset $\sS^*$ consists of optimal latent dimensions determined by an entropy-based feature selector cooperating with EC-VAE in conjunction with a Multi-Layer Perceptron (MLP) regressor $P_{\psi}$ that maps the target features to latent variables with low entropy.
In Stage 2, after embedding the target features, we transform the latent variables $\rvz$ into a uniform distribution $\tilde{\rvz}$ using a Uniform Transformation (UT) Module~\cite{shiUniformTransformationRefining2024}.
The UT module identifies the probability density function (PDF) of the latent variables and performs a transformation to achieve a uniform distribution, which ensures the feasibility of the solution and fairness during latent variable sampling.
A refined decoder $D_{\tilde{\theta}}$ is also employed to enhance the decoding quality of the uniformly transformed latent variables $\tilde{\rvz}$.
The selected latent variables $\rvz_{\sS^*}$ are then reused to construct another MLP regressor $P_{\tilde{\psi}}$, serving as a surrogate model of the original problem within the latent space.

\begin{figure}
    \centering
    \includegraphics[scale = 0.33]{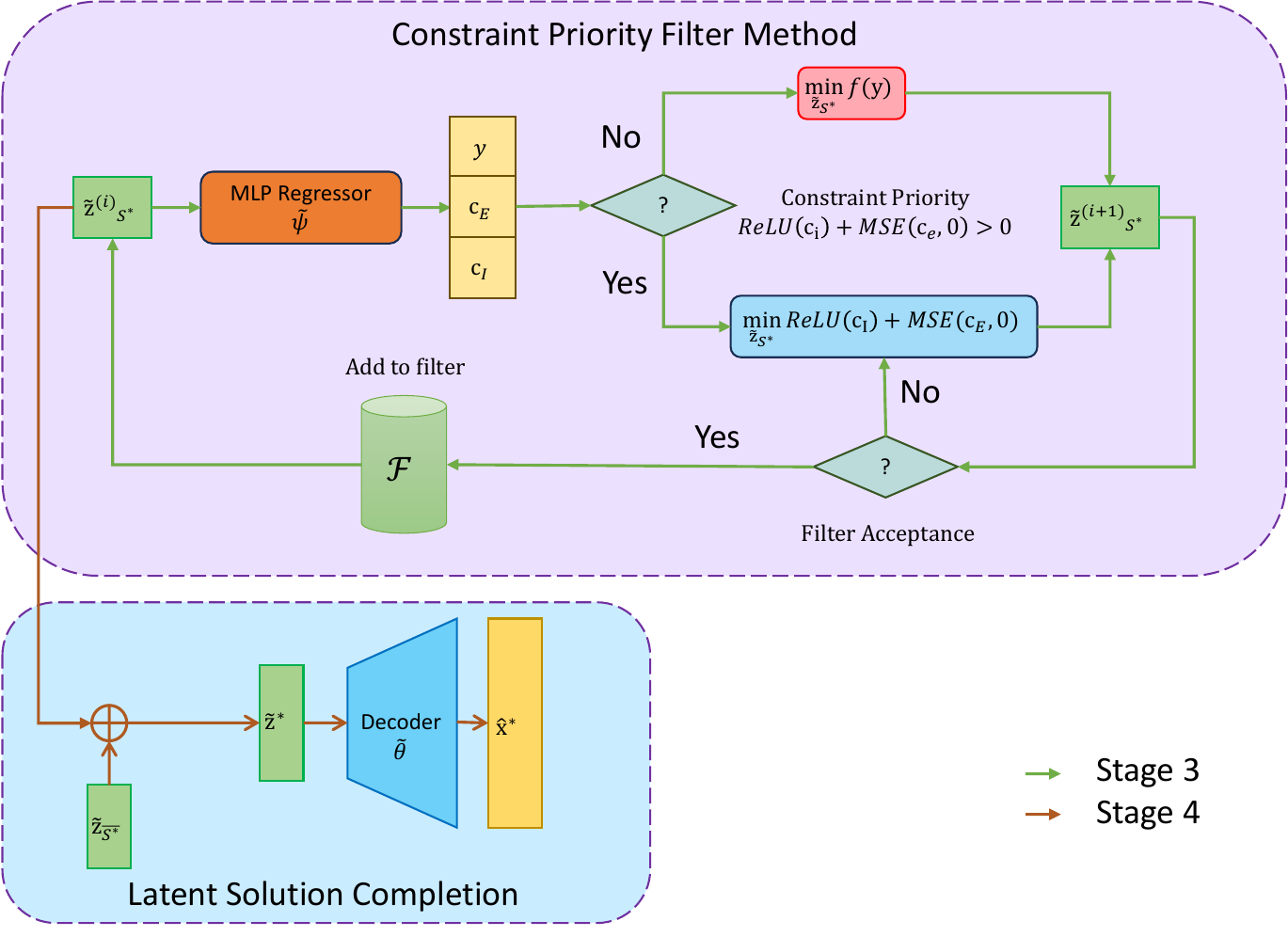}

    \caption{Diagram of Optimizing Phase.
The color-coded pipelines illustrate the processes within each stage. 
Stage 3 solves the optimization problem within the latent space. 
In Stage 4, the optimal latent solution is mapped back to the original decision variable space to generate the final solution.
}
    \label{fig:optimizing}
\end{figure}

In the \textit{optimization phase} as Fig.~\ref{fig:optimizing}, the problem is represented using the refined latent variables with selected features $\tilde{\rvz}_{\sS^*}$ through the secondary MLP regressor.
In Stage 3, we propose the CPFM which is a hard constraint optimization method to solve the problem built as a surrogate model in Stage 2.
In Stage 4, the optimal latent solution for the selected features, $\tilde{\rvz}^*_{\sS^*}$, is concatenated with randomly generated values for the unselected latent features $\tilde{\rvz}_{\bar{\sS}^*}$ to form the full optimal latent solution $\tilde{\rvz}^* \in \mathbb{R}^l$.
This latent solution is then decoded back into the original decision variable space, yielding the optimal solution $\hat{\rvx}^*$.

In summary, the challenges posed by black-box optimization problems with high-dimensional and noisy decision variables necessitate a robust and adaptive approach. 
Our MCOF transforms the problem into a latent space, enabling effective feature embedding and optimization while handling constraints. 
The subsequent sections will elaborate on the target feature embedding process, the implementation of the Uniform Transformation module alongside the refined MLP regressor, and the detailed optimizing phase of CPFM and latent solution completion.

\subsection{Target Feature Embedding}

In Stage 1, the goal is to convert the decision space into a latent space using EC-VAE. The decision space itself does not inherently encode information about the target features, necessitating the embedding of these features into the latent space.

In white-box optimization problems, researchers typically use the latent space solely as a sampling space without embedding advanced information, leveraging the VAE's well-known capabilities for data compression \cite{shwartzzivCompressNotCompress2024} and disentangled representation learning \cite{wangDisentangledRepresentationLearning2024}. However, in our case, the decision space contains interrelated features that can be leveraged to enrich the latent space and represent the target features.

To achieve this, we set up the latent space with a sufficiently large dimensionality and leverage the phenomenon of over-pruning \cite{yeungTacklingOverpruningVariational2017}. Over-pruning in VAEs is characterized by the inactivation of certain latent dimensions, leading to a sparse latent space \cite{aspertiSparsityVariationalAutoencoders2019}. 
This phenomenon acts as a double-edged sword: excessive pruning can cause posterior collapse, where encoded representations fail to carry meaningful information, resulting in poor reconstruction and generative performance. 
Conversely, sparsity in the latent space can be advantageous for disentangling and compressing representations, focusing on the most relevant features. 
Striking the right trade-off is critical for maintaining both the expressiveness and efficiency of the latent space.

To address potential encoding capacity limitations, we employ EC-VAE, which imposes a lower bound $\eta$ on the entropy of the latent variables to prevent excessive pruning. 
EC-VAE shares the same structure as the vanilla VAE, as shown in \Figr{fig:VAE}. 
The formulation of the loss is defined as follows:
\begin{equation}
\label{eqn:EC-VAE}
\begin{aligned}
    L_{\mbox{EC}} = &\; -\mathbb{E}_{q_{\phi}(\rvz | \rvx)}\left[\ln p_{\theta}(\rvx | \rvz)\right] 
    + \beta \mathrm{D}_{\mbox{KL}}\left(q_{\phi}(\rvz |\rvx) \| p(\rvz)\right) \\
    &+ \gamma \sum_{i=1}^{l}\max\left(0,\;\eta - \hat{H}(\ervz_i)\right).
\end{aligned}
\end{equation}
The first term in the loss function, $-\mathbb{E}_{q_{\phi}(\rvz | \rvx)}[\ln p_{\theta}(\rvx | \rvz)]$, represents the reconstruction loss, which ensures that the latent representation $\rvz$ retains sufficient information to accurately reconstruct the input $\rvx$. 
The second term, $\beta \mathrm{D}_{\mbox{KL}}(q_{\phi}(\rvz | \rvx) \| p(\rvz))$, regularizes the posterior distribution $q_{\phi}(\rvz|\rvx)$ to be close to the prior $p(\rvz)$, promoting a well-structured latent space. 
The third term imposes a lower bound $\eta$ on the entropy of each latent dimension, encouraging the utilization of sufficient encoding capacity and mitigating the risk of over-pruning.
All three terms are written so that $L_{\mbox{EC}}$ is minimized; the first two together form the negative of the $\beta$-weighted evidence lower bound.
The rectifier $\max(0,\cdot)$ makes the third term a genuine one-sided bound: no reward accrues for entropy beyond $\eta$, which would otherwise oppose the KL term without limit.

Here $\hat{H}(\ervz_i)$ denotes the differential entropy of the \emph{aggregate} posterior of dimension $i$,
\begin{equation}\label{eqn:aggpost}
    q_\phi(\ervz_i) \;=\; \frac{1}{N}\sum_{n=1}^{N} q_\phi(\ervz_i \mid \rvx_n),
\end{equation}
rather than that of the conditional posterior $q_\phi(\ervz_i \mid \rvx)$.
The distinction is essential. An over-pruned dimension satisfies $q_\phi(\ervz_i \mid \rvx) \approx p(\ervz_i)$ for every $\rvx$ and therefore has an aggregate entropy equal to that of the prior, whereas an active dimension spreads its code across the dataset and exceeds it.
Setting $\eta = \tfrac{1}{2}\ln(2\pi e) \approx 1.4189$, the differential entropy of $\mathcal{N}(0,1)$, thus penalizes precisely those dimensions that have collapsed to the prior.
Had $\hat{H}$ been taken over the conditional posterior, the same bound would have driven every dimension \emph{toward} collapse, since an informative dimension has $\sigma_{\ervz_i \mid \rvx} < 1$ and hence a conditional entropy below $\tfrac{1}{2}\ln(2\pi e)$.

By combining these three components, EC-VAE balances reconstruction accuracy, latent space regularization, and efficient utilization of latent dimensions.
Note that EC-VAE is intended to bound pruning rather than eliminate it: the sparsity discussed above remains available for compressing the representation, while the entropy floor prevents it from degenerating into posterior collapse.

In the EC-VAE, latent dimensions with lower entropy tend to correspond to redundant or irrelevant information in the data, while higher entropy dimensions embed meaningful underlying factors. 
To embed the target features into the latent space, we deliberately use the lower-entropy latent dimensions. 
This approach ensures that the relevant features are preserved for other optimization purposes, while the target features are embedded in the dimensions that encode redundant or less relevant information.

The target features---comprising the objective value $y$, equality constraints $\rvc_E$, and inequality constraints $\rvc_I$---are embedded into a subset of the latent variables $\rvz_{\sS^*}$ through an MLP regressor:
$$
\begin{bmatrix}
    y & \rvc_E & \rvc_I
\end{bmatrix}^\mathsf{T} 
\,=\, P_{\psi}\bigl(\rvz_{\sS}\bigr),
$$
where $P_{\psi}$ is trained to minimize a regression loss $L_{\mathrm{Reg}}$, and $\rvz_{\sS}$ denotes the selected latent dimensions. 
To maintain a fixed network architecture, we introduce a latent feature selector that operates as
\begin{equation}\label{eqn:selector}
    \rvz_{\sS} \;=\; \rvw \odot \rvz,
    \qquad
    \rvw \in \{0,1\}^{l},
    \qquad
    \sS = \{\, i : \ervw_i = 1 \,\},
\end{equation}
where $\odot$ denotes the Hadamard (elementwise) product and the binary mask $\rvw$ specifies which latent dimensions are used.
Masking rather than sub-selecting keeps the regressor's input width fixed at $l$ as $|\sS|$ shrinks.
The mask is binary rather than $1/|\sS|$-valued so that pruning does not rescale the regressor's input, which would otherwise shift the effective value of the threshold $\alpha$ used in Algorithm~\ref{alg:EC-VAE} each time a dimension is removed.
The loss function $L_{\mathrm{Reg}}$ is chosen according to the nature of the target features: mean squared error (MSE) for continuous values and cross-entropy for categorical values, ensuring that the embedded features are accurately captured within the lower-entropy latent dimensions.

To incorporate both the EC-VAE and the regressor into a unified training procedure, we minimize the two objectives jointly:
\begin{equation}\label{eqn:joint}
    \min_{\phi,\,\theta,\,\psi}\,\Bigl[
    L_{\mathrm{EC}}\bigl(\phi,\theta;\,\rvx\bigr)
    \;+\;
    \lambda\,L_{\mathrm{Reg}}\bigl(\psi;\,\rvw \odot E_\phi(\rvx)\bigr)
    \Bigr],
\end{equation}
where $E_\phi(\rvx)$ is the encoder and $\lambda > 0$ balances the two terms.
Algorithm~\ref{alg:EC-VAE} iteratively optimizes the EC-VAE parameters $(\phi,\theta)$ while refining the feature selector mask $\rvw$. 
Initially every dimension is selected, so that $\rvw = \mathbf{1}_l$ and $s = l$. 
During each training iteration, the feature selector and the EC-VAE update their parameters based on the latent dimension entropies, giving priority to those with lower entropy for embedding the target features. 
Simultaneously, the MLP regressor $P_{\psi}$ is optimized to minimize errors in reconstructing the target outputs. 
When the validation loss of the regressor exceeds a predefined threshold $\alpha$, the highest-entropy dimensions currently in $\sS$ are removed and the mask is reassigned to the $s$ lowest-entropy dimensions.
The rationale is that the high-entropy dimensions encode data features rather than target features, so that their presence in $\sS$ acts as noise for $P_\psi$; removing them concentrates the target embedding on the redundant capacity where it belongs.
The subset size $s$ is persistent state carried across epochs, is shrunk geometrically at rate $\rho$ so that a wide latent space can be reduced within a practical number of epochs, and is floored at $s_{\min}$.
Training terminates once the regressor has held its accuracy on an unchanged subset for $T$ consecutive epochs, ultimately yielding a compact and expressive latent space where the critical objective and constraint information is embedded in lower-entropy dimensions.


\begin{algorithm}[htb]
  \caption{Training Procedure of EC-VAE to Embed Target Features}
  \label{alg:EC-VAE}
  \begin{algorithmic}[1]
  \renewcommand{\algorithmicrequire}{\textbf{Input:}}
  \renewcommand{\algorithmicensure}{\textbf{Output:}}
  \REQUIRE Decision variable set $\sX$; target set $\{\sY, \sC_E, \sC_I\}$; latent width $l$; regressor validation loss threshold $\alpha$; pruning rate $\rho \in (0,1)$; minimum subset size $s_{\min}$; patience $T$; epoch budget $E_{\max}$
  \ENSURE Feature selector mask $\rvw$, selected set $\sS^*$, EC-VAE model parameters $\theta$ and $\phi$, MLP regressor parameters $\psi$
  \STATE Select all dimensions initially and record the subset size:
  $$
  \ervw_i \gets 1 \;\; \text{for} \;\; i = 1, \dots, l,
  \qquad s \gets l, \qquad t \gets 0.
  $$
  \FOR{$e = 1$ to $E_{\max}$}
    \FOR{each batch}
      \STATE Update $\psi$ using the loss function $L_{\mathrm{Reg}}$
      \STATE Update $\theta$ and $\phi$ using the combined loss $L_{\mathrm{EC}} + \lambda L_{\mathrm{Reg}}$
    \ENDFOR
    \STATE Evaluate the regressor validation loss $a$
    \IF{$a > \alpha$ \AND $s > s_{\min}$}
      \STATE Estimate the aggregate-posterior entropies of all $l$ dimensions:
      $$\sH = \{\hat{H}(\ervz_1), \hat{H}(\ervz_2), \dots, \hat{H}(\ervz_l)\}$$
      \STATE Sort indices in ascending order of entropy:
      $$
      \rvr \gets \text{argsort}(\sH),
      $$
      where $\rvr = \begin{bmatrix}
          \ervr_1 & \ervr_2 & \cdots & \ervr_l
      \end{bmatrix}^\mathsf{T}$ and $\ervr_i$ is the dimension index at the $i$th order.
      \STATE Shrink the subset, subject to the floor:
      $$
      s \gets \max\bigl(s_{\min},\; \lceil (1-\rho)\,s \rceil\bigr).
      $$
      \STATE Retain the $s$ lowest-entropy dimensions:
      $$
      \rvw \gets \mathbf{0}, \qquad \ervw_{\ervr_i} \gets 1 \;\; \text{for} \;\; i = 1, \dots, s.
      $$
      \STATE Reset the patience counter: $t \gets 0$
    \ELSE
      \STATE $t \gets t + 1$
      \STATE \textbf{if} $t \geq T$ \textbf{then break}
    \ENDIF
  \ENDFOR
  \STATE $\sS^* \gets \{\ervr_1, \ervr_2, \dots, \ervr_s\}$
  \end{algorithmic}
\end{algorithm}

On termination, $\sS^* = \{\ervr_1, \ervr_2, \cdots, \ervr_s \}$ is the selected set of embedding latent dimensions, and $\rvw$ is the corresponding binary mask.

\subsection{Uniform Transformation to Refine the Decoder and MLP Regressor}

\begin{figure}[htb]
    \centering
    \includegraphics[width=0.9\columnwidth]{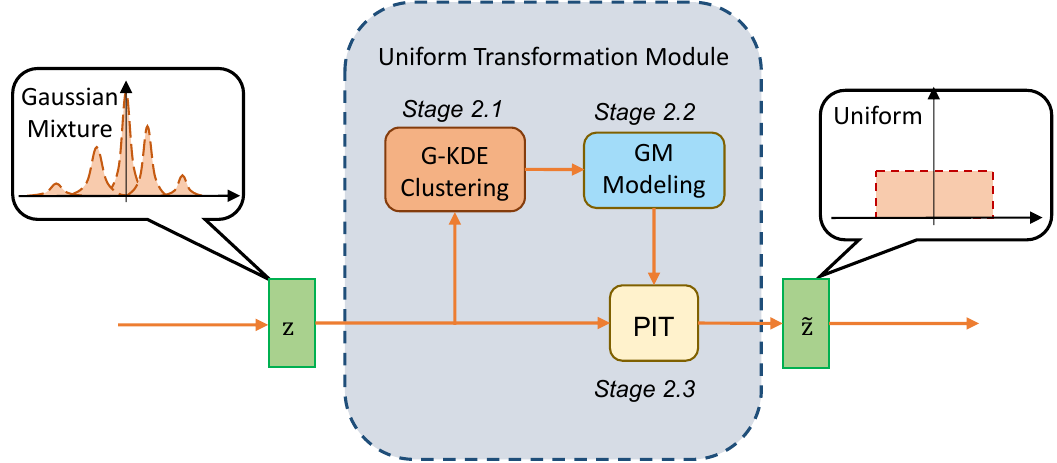}
    \caption{Schematic of the Uniform Transformation Module. The module, highlighted within the gray box, transforms latent variables $\rvz$ to $\tilde{\rvz}$, representing the pre- and post-transformation states, respectively.}
    \label{fig:UTModule}
\end{figure}

In Stage 1, we built an MLP regressor $P_\psi$ for the target variables to construct the original problem in the latent space. However, in deep learning-based inverse problems, researchers often encounter invalid solutions due to the increasing trend of over-parameterization---"the deeper, the better." This trend expands the decision space of optimization problems, exacerbating non-convexity and introducing numerous local optima.

In the context of VAE optimization, data features are embedded in a constrained latent space and represented as a Gaussian Mixture (GM) distribution that often cannot fully capture the original data distribution. This limitation results in biases when downstream models are built, as they are trained on over-represented or skewed data features. This bias is akin to an imbalanced training data problem, where certain data regions dominate the learning process.

To address these challenges, we implement the Uniform Transformation (UT) module~\cite{shiUniformTransformationRefining2024} in Stage 2 to refine latent representations, enhance the decoder, and support the construction of a robust MLP regressor. The UT module is specifically designed to transform GM-distributed latent variables $\rvz$ into a uniform distribution $\tilde{\rvz}$. By transforming GM latent variable distributions into a uniform distribution, the UT module addresses the critical discrepancies caused by the mismatch between the encoder and decoder distributions in VAEs~\cite{koike-akinoAutoVAEMismatchedVariational2022}.

The UT module operates in three sub-stages, as illustrated in Fig.~\ref{fig:UTModule} and detailed below.

\subsubsection*{Stage 2.1: Gaussian Kernel Density Estimation (G-KDE) Clustering Algorithm}
\begin{figure}[htb]
    \centering
    \includegraphics[width=0.6\linewidth]{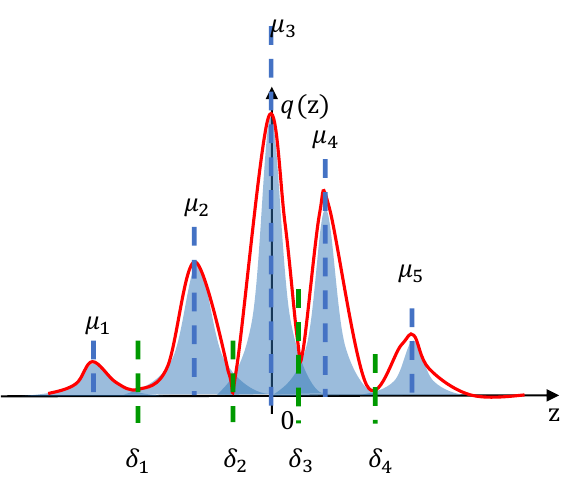}
    \caption{Histogram of Gaussian Mixture Distribution of Latent Variable. The red curve represents the estimated PDF by G-KDE. The green dashed lines are the thresholds detecting the child Gaussians. The blue dashed lines represent the centroids of the child Gaussians.}
    \label{fig:GM}
\end{figure}

We apply the G-KDE clustering algorithm to accurately identify the child Gaussian distributions within the latent-variable distribution and obtain the Gaussian Mixture (GM) model parameters as shown in \Figr{fig:GM}.

The G-KDE estimates the probability density function (PDF) of the latent variable $z$ using:

\begin{equation}\label{eqn:GKDE}
\hat{f}(z) = \frac{1}{n h} \sum_{i=1}^{n} \mathcal{K}\left( \frac{z - z_i}{h} \right),
\end{equation}

where $\hat{f}(z)$ denotes the estimated PDF, $n$ is the number of data points, $z_i$ are the observed data samples of a single latent dimension, $h$ is the bandwidth set using the Scott method as $h = n^{-1/5}\sigma$, and $\mathcal{K}(\cdot)$ is the Gaussian kernel function $\mathcal{K}(u) = \frac{1}{\sqrt{2\pi}} e^{-u^2/2}$.

After estimating the PDF, we identify the local maxima and minima to determine the means $\{\hat{\mu}^{(k)}\}$ and cluster thresholds $\{\delta_j\}$ of the child Gaussians in the GM distribution. The variance $\left(\hat{\sigma}^{(k)}\right)^2$ and weight $\hat{\omega}^{(k)}$ of each child Gaussian are computed based on the respective cluster's data.

The process of Stage 2.1 is outlined in Algorithm~\ref{alg:GKDE}.

\begin{algorithm}[htb!]
\caption{G-KDE Clustering Algorithm for a Single Latent Dimension}\label{alg:GKDE}
\begin{algorithmic}[1]
  \renewcommand{\algorithmicrequire}{\textbf{Input:}}
  \renewcommand{\algorithmicensure}{\textbf{Output:}}
    \REQUIRE $\sZ = \{z_1, z_2, \dots, z_n\}$: array of latent variable samples for a single latent dimension.
    \ENSURE Cluster labels for each $z_i$, centroids $\{\hat{\mu}^{(k)}\}$, variances $\{\left(\hat{\sigma}^{(k)}\right)^2\}$, weights $\{\hat{\omega}^{(k)}\}$.
    \STATE Estimate the PDF $\hat{f}(z)$ of $\sZ$ using G-KDE as in \eqref{eqn:GKDE}.
    \STATE Identify local maxima $\{\hat{\mu}^{(k)}\}$ and minima $\{\delta_j\}$ of $\hat{f}(z)$.
    \FOR{each $z_i$}
        \IF{$z_i \leq \delta_1$}
            \STATE Assign $z_i$ to Cluster $1$.
        \ELSIF{$\delta_j < z_i \leq \delta_{j+1}$ \textbf{for} $j = 1$ to $K-1$}
            \STATE Assign $z_i$ to Cluster $j+1$.
        \ELSIF{$z_i > \delta_{K-1}$}
            \STATE Assign $z_i$ to Cluster $K$.
        \ENDIF
    \ENDFOR
    \FOR{each Cluster $k$}
        \STATE Compute variance:
        $$
        \left(\hat{\sigma}^{(k)}\right)^2 = \frac{1}{|\sZ^{(k)}|-1} \sum_{z \in \sZ^{(k)}} (z - \hat{\mu}^{(k)})^2,
        $$
        where $|\sZ^{(k)}|$ is the number of samples in Cluster $k$.
        \STATE Compute weight:
        $$
        \hat{\omega}^{(k)} = \frac{|\sZ^{(k)}|}{n}.
        $$
    \ENDFOR
\end{algorithmic}
\end{algorithm}

\subsubsection*{Stage 2.2: Modeling the GM Distribution}

We reconstruct the posterior GM distribution of the latent variable $z$ using the parameters obtained from Stage 2.1:

\begin{equation}\label{eqn:GMPDF}
\hat{q}_{\phi}(z| \rvx) = \sum_{k=1}^{K} \hat{\omega}^{(k)} \hat{q}_{k}\left(z\right),
\end{equation}

where $\hat{q}_{k}\left(z\right) = \mathcal{N}\left(\hat{\mu}^{(k)}, \left(\hat{\sigma}^{(k)}\right)^2\right)$ represents the PDF of each child Gaussian, and $\hat{\omega}^{(k)}$ are the corresponding weights.

\subsubsection*{Stage 2.3: Probability Integral Transform (PIT) of Latent Variable}

We apply the Probability Integral Transform (PIT)~\cite{davidProbabilityIntegralTransformation1948} to transform the irregular GM distribution into a uniform distribution. The PIT is defined as:

\begin{equation}
\tilde{z} = \sum_{k=1}^{K} \hat{\omega}^{(k)} \hat{Q}_{k}\left(z\right),
\end{equation}

where $\hat{Q}_{k}\left(z\right)$ is the cumulative distribution function (CDF) of the $k$-th child Gaussian $\hat{q}_{k}\left(z\right)$. This transformation ensures that $\tilde{z}$ follows a uniform distribution over $[0, 1]$. We scale $\tilde{z}$ to the range $[-4, 4]$ to prevent gradient vanishing during training.

\subsubsection*{Retraining the Decoder and MLP Regressor}

Since the latent representations have now been transformed to a uniform distribution, we need to retrain the decoder and the MLP regressor with new parameters $\tilde{\theta}$ and $\tilde{\psi}$. The new MLP regressor $P_{\tilde{\psi}}$ uses the optimal embedded latent dimensions $\sS^*$ from the feature selector to predict the target features. The new decoder $D_{\tilde{\theta}}$ decodes the refined latent variables $\tilde{\rvz}$ to the original decision variable $\rvx$.

This completes the training phase of our framework, transferring the original problem from the decision space $\sX$ to the refined latent space $\tilde{\sZ}$.

\subsection{Constraint-priority Filter Method}

In the training phase, we represent the optimization problem in the latent space, simplifying it as follows:
\begin{equation}
\label{eq:representedProblem}
\begin{aligned}
    \min_\zeta &\quad \hat{y}(\zeta), \\
    \text{subject to}  &\quad \hat{\rvc}_E(\zeta) = \mathbf{0},\\
    &\quad \hat{\rvc}_I(\zeta) \leq \mathbf{0},     
\end{aligned}
\end{equation}
where $\zeta = \tilde{\rvz}_{\sS^*}$ represents the selected refined latent variables, and
\[
\begin{bmatrix}
    \hat{y}(\zeta) & \hat{\rvc}_E(\zeta) & \hat{\rvc}_I(\zeta) 
\end{bmatrix}^\mathsf{T} = P_{\tilde{\psi}} (\zeta)
\]
denotes the predicted values of the target features, including the objective $\hat{y}(\zeta)$ and constraints $\hat{\rvc}_I(\zeta)$ (inequality) and $\hat{\rvc}_E(\zeta)$ (equality).

Instead of relying on the Lagrangian method, which requires searching for multipliers~\cite{nandwaniPrimalDualFormulation2019,fiorettoLagrangianDualityConstrained2021, parkSelfSupervisedPrimalDualLearning2023} and often leads to instability in back-propagation~\cite{marquez-neilaImposingHardConstraints2017}, 
we propose a constraint-priority filter method (CPFM) to address constrained deep learning optimization in Stage 3. 
This approach improves the stability and robustness of the optimization process by avoiding the challenges associated with the estimation of multipliers. 

Following the filter method paradigm, we reframe the problem as a bi-objective optimization:
\begin{equation}\label{eqn:biobjective}
    \min_\zeta \Bigl(o(\zeta),\,v(\zeta)\Bigr),
\end{equation}
where \(o(\zeta) = \hat{y}(\zeta)\) is the objective function and
\begin{equation}\label{eqn:violation}
    v(\zeta) \;=\; \bigl\|\hat{\rvc}_E(\zeta)\bigr\|_2^{2} \;+\;\bigl\|\max\bigl(\mathbf{0},\;\hat{\rvc}_I(\zeta)\bigr)\bigr\|_2^{2}
\end{equation}
quantifies constraint violation, where $\max(\cdot,\cdot)$ acts elementwise so that both terms reduce to scalars for vector-valued constraints. 
Equality constraints are captured by a squared \(\ell_2\)-norm, while inequality constraints use a rectified squared term to penalize positive violations. 
The norms are squared so that $v$ is differentiable everywhere, including at feasible points: the unsquared \(\ell_2\)-norm has a non-vanishing gradient exactly at $\hat{\rvc}_E = \mathbf{0}$, which is where the violation-reduction subproblem is driven, and would reintroduce the very oscillation near feasibility that motivates avoiding multiplier-based methods. 
By construction $v(\zeta) \geq 0$, with equality if and only if $\zeta$ is feasible for the surrogate, and $\nabla v(\zeta) = \mathbf{0}$ at every such point. 
This reformulation thus enables us to prioritize constraint satisfaction and objective improvement concurrently.

We emphasize that $v$ is built from the learned regressor $P_{\tilde\psi}$, so the feasibility CPFM certifies is feasibility \emph{with respect to the surrogate}. 
The gap to the true black-box constraints is bounded by the surrogate error and is quantified empirically in Section~\ref{sec:experiment}.

\begin{algorithm}[htb!]
\caption{Constraint-priority Filter Method}
\label{alg:CPFM}
\begin{algorithmic}[1]
    \REQUIRE Initial solution $\zeta^{(0)}$; tolerances $\epsilon_v$, $\epsilon_o$; envelope parameters $\gamma_v, \gamma_o \in (0,1)$; initial radius $\Delta_0$; minimum radius $\Delta_{\min}$; contraction factor $\tau \in (0,1)$; inner budget $K$; maximum iterations $N_{\text{max}}$.
    \ENSURE Optimized solution $\zeta^*$, updated filter $\mathcal{F}$.
    
    \STATE Initialize the filter with the starting pair:
    $$\mathcal{F} \gets \{(v(\zeta^{(0)}),\, o(\zeta^{(0)}))\}.$$
    \STATE Set $\Delta \gets \Delta_0$ and $i \gets 0$.
    
    \WHILE{$i < N_{\text{max}}$}
        \IF{$v(\zeta^{(i)}) > \epsilon_v$}
            \STATE Obtain $\zeta^{+}$ by $K$ projected-gradient steps on $\min_\zeta v(\zeta)$ from $\zeta^{(i)}$, restricted to $\|\zeta - \zeta^{(i)}\|_\infty \leq \Delta$.
        \ELSE
            \STATE Obtain $\zeta^{+}$ by $K$ projected-gradient steps on $\min_\zeta o(\zeta)$ from $\zeta^{(i)}$, restricted to $\|\zeta - \zeta^{(i)}\|_\infty \leq \Delta$.
        \ENDIF
        \STATE Compute $v(\zeta^{+})$ and $o(\zeta^{+})$.
        
        \STATE \textbf{Filter acceptance:} $\zeta^{+}$ is acceptable to $\mathcal{F}$ if, for \emph{every} $(v_j, o_j) \in \mathcal{F}$,
        \begin{equation*}
            v(\zeta^{+}) \leq (1-\gamma_v)\,v_j
            \quad \text{or} \quad
            o(\zeta^{+}) \leq o_j - \gamma_o\,v_j .
        \end{equation*}
        
        \IF{$\zeta^{+}$ is acceptable}
            \STATE $\zeta^{(i+1)} \gets \zeta^{+}$ and $\Delta \gets \Delta_0$.
            \STATE Add the new pair and prune the filter:
            $$\mathcal{F} \gets \bigl(\mathcal{F} \cup \{(v(\zeta^{+}), o(\zeta^{+}))\}\bigr) \setminus \mathcal{D},$$
            where $\mathcal{D}$ collects the pairs dominated by $(v(\zeta^{+}), o(\zeta^{+}))$.
            \IF{$v(\zeta^{(i+1)}) \leq \epsilon_v$ \AND $|o(\zeta^{(i+1)}) - o(\zeta^{(i)})| \leq \epsilon_o$ \AND the last step minimized $o$}
                \STATE \textbf{Break loop}.
            \ENDIF
            \STATE Update $i \gets i + 1$.
        \ELSE
            \STATE Contract the trust region and retry from $\zeta^{(i)}$: $\Delta \gets \tau \Delta$.
            \STATE \textbf{if} $\Delta < \Delta_{\min}$ \textbf{then break}
        \ENDIF
    \ENDWHILE
    
    \IF{$v(\zeta^{(i)}) \leq \epsilon_v$}
        \STATE \textbf{Return} $\zeta^* \gets \zeta^{(i)}$, updated filter $\mathcal{F}$.
    \ELSE
        \STATE \textbf{Return} failure (iteration budget or minimum radius reached).
    \ENDIF
\end{algorithmic}
\end{algorithm}

Our proposed CPFM is detailed in \Algref{alg:CPFM}.
CPFM begins by initializing the filter $\mathcal{F}$, which stores pairs of constraint violations and objective values, with the pair obtained from the starting point $\zeta^{(0)}$.
The algorithm then iteratively solves subproblems based on the current state of constraint violations by inverse optimization: 
\begin{itemize}
    \item if $v(\zeta^{(i)}) > \epsilon_v$, the focus is on minimizing $v(\zeta)$ to reduce constraint violations; 
    \item otherwise, when $v(\zeta^{(i)}) \leq \epsilon_v$, the emphasis shifts to optimizing the primary objective $o(\zeta)$. 
\end{itemize}
This alternation is what we mean by \emph{constraint priority}: an objective step is taken only from a point that is already feasible to tolerance.
Since our problems are represented by neural networks, gradient-based optimization using back-propagation is applied to solve these subproblems, restricted at each iteration to a trust region of radius $\Delta$ so that the step length remains controllable.

A candidate is accepted when it is \emph{not dominated} by any entry of the filter, that is, when for every stored pair it improves either the violation or the objective.
Following classic filter methods~\cite{fletcher_brief_2006}, acceptance additionally requires a sufficient reduction relative to each entry, governed by the envelope parameters $\gamma_v$ and $\gamma_o$.
The envelope matters: without it the filter admits steps whose improvement vanishes in the limit, and the iterates may accumulate at an infeasible point.
Accepted pairs are added to $\mathcal{F}$ and any entries they dominate are discarded, so that the filter retains an approximation of the Pareto frontier rather than a single point.
When a candidate is rejected, the trust region is contracted by the factor $\tau$ and the step is retried from the same iterate; this is the mechanism by which progress is recovered, and it terminates once $\Delta$ falls below $\Delta_{\min}$.

The algorithm checks for convergence after each accepted step, terminating when the violation $v(\zeta^{(i+1)})$ falls below the tolerance $\epsilon_v$ and the objective shows no significant further improvement.
The objective test is applied only when the preceding step minimized $o$, since consecutive iterates produced by different subproblems carry no information about stationarity of the objective.
Upon termination, the algorithm returns the optimized solution $\zeta^*$ and the final filter, which encapsulates the trade-offs explored during optimization.

\subsection{Latent Solution Completion}
In Stage 4, we have obtained the solution of the transferred problem in \Eqr{eq:representedProblem}, $\tilde{\rvz}_{\sS^*} = \zeta^*$, where $\sS^*$ is the set of selected embedding dimensions in the latent space. 
Therefore, when $|\sS^*| < l$, meaning that the embedding dimensions do not cover the entire latent space, the remaining latent dimensions $\bar{\sS^*}$ contain irrelevant variables to the original problem, denoted as $\tilde{\rvz}_{\bar{\sS^*}}$. 

To complete the latent representation from the solution, we sample the irrelevant variables as follows:
\begin{equation}
\tilde{\ervz}^*_i =
\begin{cases} 
    \zeta^*_i, & \text{if } i \in \sS^*, \\
    \tilde{\ervz}_i \sim \mathcal{U}(-4, 4), & \text{if } i \in \bar{\sS^*},
\end{cases}
\end{equation}
where $\tilde{\ervz}^*_i$ represents the $i$-th dimension of the completed latent solution $\tilde{\rvz}^*$. 
For dimensions $i \in \sS^*$, the values are taken directly from the solution $\tilde{\rvz}_{\sS^*}$ obtained from the optimization process. 
For dimensions $i \in \bar{\sS^*}$, the irrelevant latent variables are sampled uniformly from $\mathcal{U}(-4, 4)$. 
The range $[-4, 4]$ is chosen to prevent gradient vanishing or exploding during subsequent decoding or inference steps at the UT module.

This completed latent solution $\tilde{\rvz}^*$ is then passed to the decoder $D_{\tilde{\theta}}$, which reconstructs the corresponding decision variable $\hat{\rvx}^* = D_{\tilde{\theta}}(\tilde{\rvz}^*)$ in the original decision space.

\section{Experimental Results}\label{sec:experiment}
We demonstrate the effectiveness of our framework in solving data-driven constrained optimization problems through two case studies: a synthetic problem and a drug discovery problem.
In the first case study, a synthetic problem is used to perform ablation analysis, systematically validating the functionality of each stage in the framework.
In the second case study, a drug discovery problem is presented to showcase the framework's performance and capability in addressing large-scale, mixed-integer optimization challenges.

\subsection{Synthetic Problem Setup}

To evaluate the functionality of our framework, we set up a synthetic problem with noisy data. We generate the problem by sampling 10,000 points of $\rvx \in \sR^{10}$, where $\rvx \sim \mathcal{U}(-50, 50)$. The optimization problem is defined as follows:
\begin{equation}\label{eqn:problem1}
\begin{aligned}
   \min_{\rvx \in \sR^{10}} \quad y &= \ervx_1^2 + \ervx_2^2,\\
   \text{subject to} \quad c_E & = \ervx_3 - \ervx_1 - 10 = 0,\\
   c_I & = 8 - (\ervx_2 - \ervx_1) \leq 0,
\end{aligned}
\end{equation}
where $y$ is the objective value to minimize, $c_E$ is the equality constraint, and $c_I$ is the inequality constraint.

This problem is treated as a data-driven black-box optimization task in our framework. Specifically, the solver does not have access to the explicit form of \Eqr{eqn:problem1}. 
Instead, only the dataset $\{\sX, \sY, \sC_E, \sC_I\}$, derived from the problem, is available for training and optimization.

The EC-VAE shares the same structure in \Figr{fig:VAE}.
Both the encoder and decoder are three-layer MLPs with leaky ReLU activations (negative slope $0.2$), maintaining a relatively minimal deep learning model. 
The latent dimension is chosen to be $10$, matching the dimensionality of the input $\rvx$, where we leverage over-pruning to reduce irrelevant latent variables. 
The encoder transforms the 10-dimensional input space into a more compact representation through two hidden layers of dimensions 256 and 128; 
the decoder performs the inverse mapping, reconstructing the input from the latent representation. 
An additional MLP regressor maps the 10-dimensional latent space to the three target features ($y, \ervc_E, \ervc_I$) using one hidden layer of dimension 1,000. 
Both the refined decoder and the refined MLP used in subsequent stages share the same network dimensions.

In the EC-VAE, we select the multipliers as $\beta = 6$, $\gamma = 1$ and $\eta = \tfrac{1}{2}\ln(2\pi e) \approx 1.42$, the differential entropy of $\mathcal{N}(0,1)$.
The tolerances in Alg.~\ref{alg:CPFM} are set as $\epsilon_o = \epsilon_v = 1\times 10^{-5}$. 

\subsection{Results of the Synthetic Problem}

\begin{table*}[htb]
\centering
\caption{Statistics of 1,000 Sampled Solutions.}
\label{tbl:x_stats}
\begin{tabular}{lc*{10}{c}} 
\hline
 & $\ervx_1$ & $\ervx_2$ & $\ervx_3$ & $\ervx_4$ & $\ervx_5$ & $\ervx_6$ & $\ervx_7$ & $\ervx_8$ & $\ervx_9$ & $\ervx_{10}$ \\
\hline
\textbf{Mean} & $-4.001$  & $4.006$   & $6.002$   & $16.66$   & $-7.223$  & $0.9968$  & $-4.93$   & $0.0178$  & $-0.9256$ & $2.033$   \\
\textbf{Std.} & $0.0036$  & $0.0034$  & $0.0037$  & $23.21$   & $45.82$   & $41.06$   & $45.81$   & $49.66$   & $67.22$   & $42.73$   \\
\hline
\end{tabular}
\end{table*}
We sampled 1,000 solutions in Stage 4 using the optimal $\zeta^*$, with statistics provided in Table~\ref{tbl:x_stats}.
\begin{table}[htb]
\centering
\caption{Statistics of Target Feature Values of 1,000 Sampled Solutions.}
\label{tbl:y_stats}
\begin{tabular}{lccc}
\hline
             & $y$                  & $c_E$                        & $c_I$                         \\ \hline
\textbf{Optimal}     & $32$                 & $0$                          & $0$                           \\
\textbf{Estimation}  & $32.0 \pm 8.0\text{e-07}$ & $-4.4\text{e-3} \pm 3.3\text{e-08}$  & $-0.036 \pm 2.6\text{e-06}$ \\
\textbf{Validation}  & $32.06 \pm 0.011$  & $2.7\text{e-3} \pm 4.2\text{e-3}$  & $-7.6\text{e-3} \pm 1.4\text{e-3}$ \\ \hline
\end{tabular}
\end{table}
Table~\ref{tbl:y_stats} compares the target feature values among the optimal solution, estimated values from the refined regressor $P_{\tilde{\psi}} (\zeta^*)$, and validation values obtained by calculating from the sampled solutions.

The results demonstrate that the sampled solutions closely match the expected target values, confirming the effectiveness of the refined regressor. 
Furthermore, the constraint values $c_E$ and $c_I$ remain close to zero, indicating that the sampled solutions effectively satisfy the problem constraints.

\subsection{Ablation Study of the Framework}
In this ablation study, we evaluate the contribution of each stage, except Stage 4 which is a solution completion, by comparing the model output with and without the proposed methods.

\subsubsection{EC-VAE and Feature Selector}
\begin{figure}[htb]
    \centering
    \subfloat[3D Scatter Plot of Latent Variables and Objective Value in VAE]{%
        \includegraphics[width=0.45\linewidth]{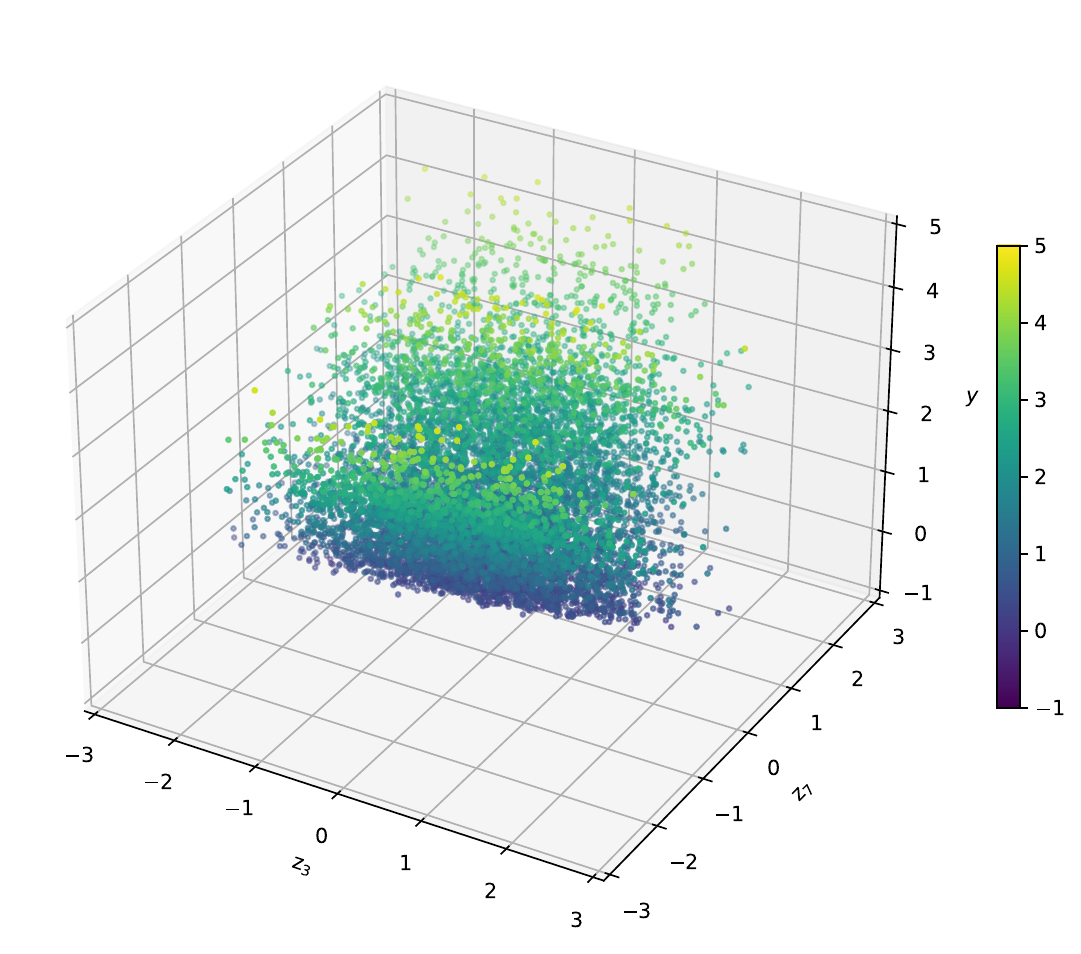}
        \label{fig:s13dvae}}
    \hfil
    \subfloat[3D Scatter Plot of Latent Variables and Objective Value in EC-VAE]{%
        \includegraphics[width=0.45\linewidth]{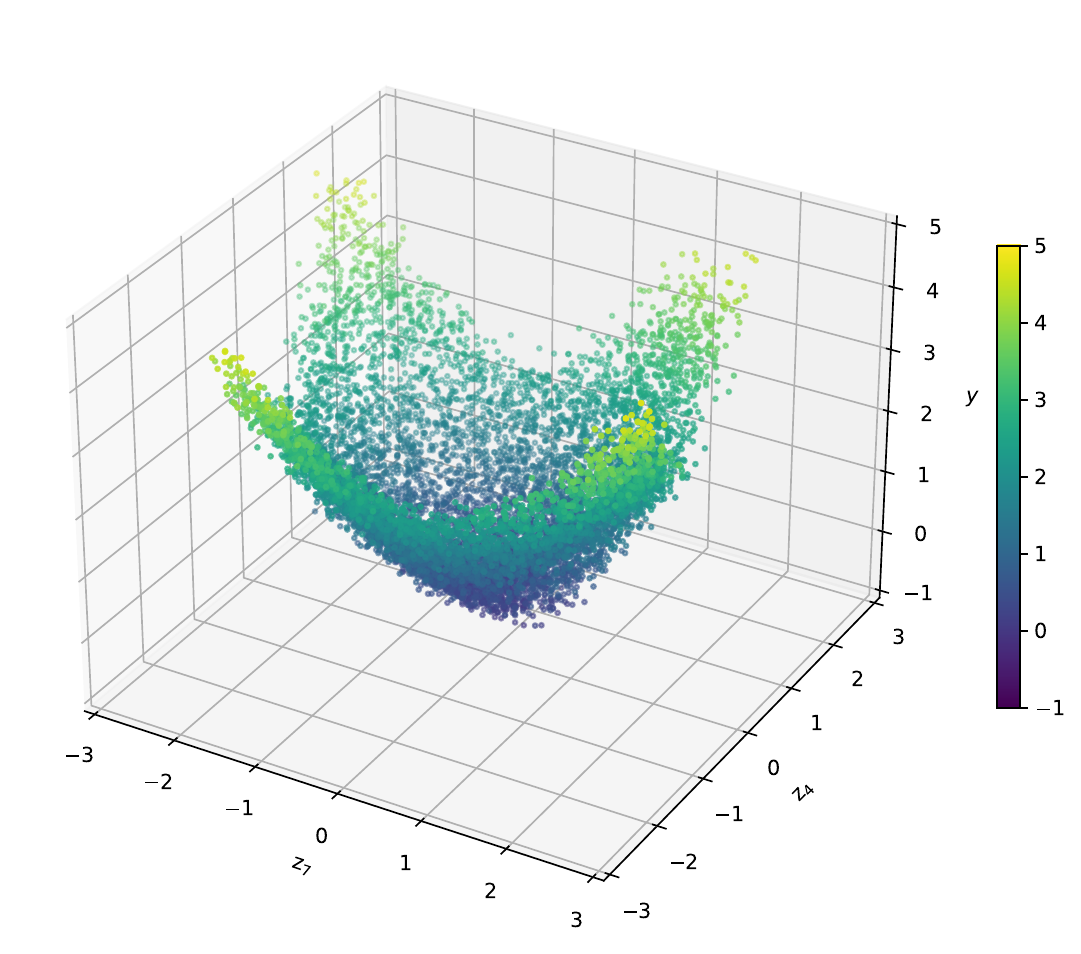}%
        \label{fig:s13dec}}
    \caption{Comparison of decision variable representations in the selected latent space for VAE and EC-VAE.
    The x-axis and y-axis represent the latent variables with the two lowest entropies, while the z-axis corresponds to the objective value from the dataset.}
    \label{fig:commapping}
\end{figure}

Figure~\ref{fig:commapping} compares the latent feature embeddings produced by VAE (without entropy constraints) and EC-VAE. 
The feature selector in Stage 1 embeds the target features into latent variables with the lowest entropy. 
EC-VAE demonstrates improved capture of decision variables compared to VAE, highlighting the effectiveness of the feature selector.

\begin{figure}[htb]
    \centering
    \subfloat[Latent Variable Distributions in VAE]{%
        \includegraphics[width=\linewidth]{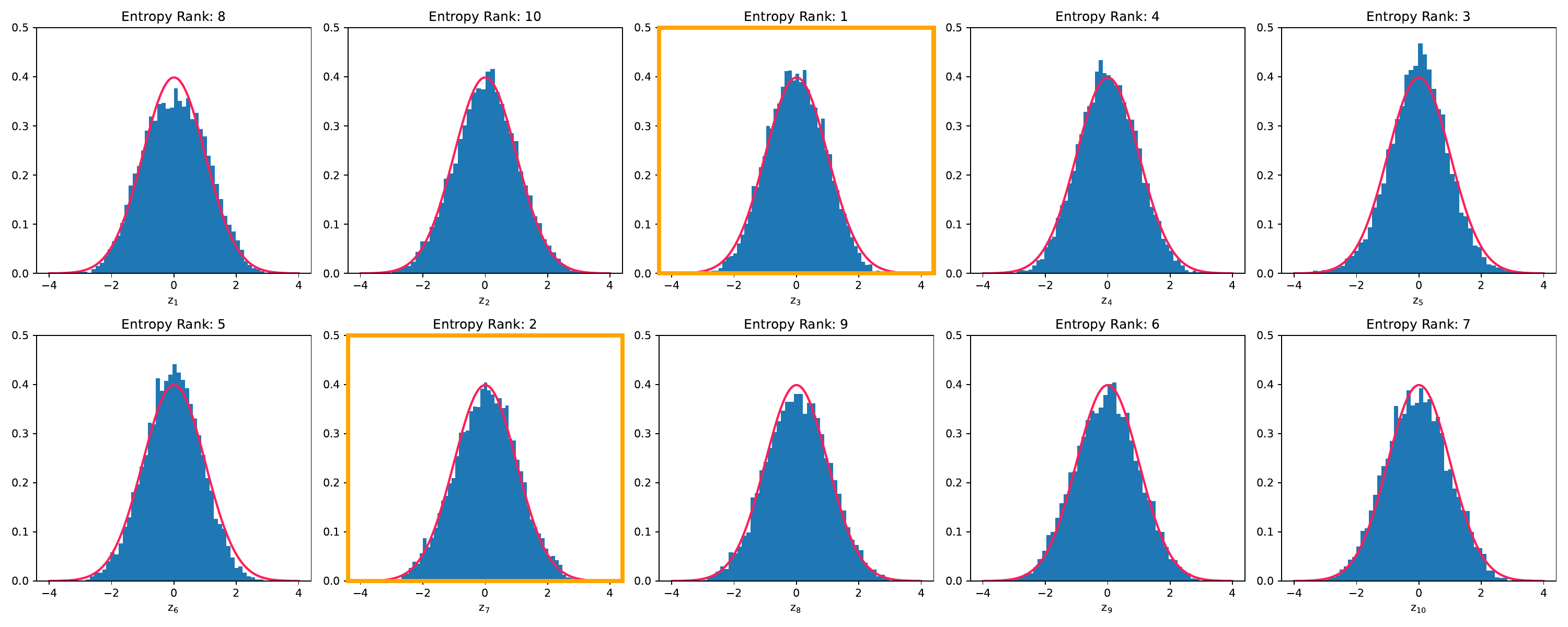}%
        \label{fig:histvae}}
    \vfil
    \subfloat[Latent Variable Distributions in EC-VAE]{%
        \includegraphics[width=\linewidth]{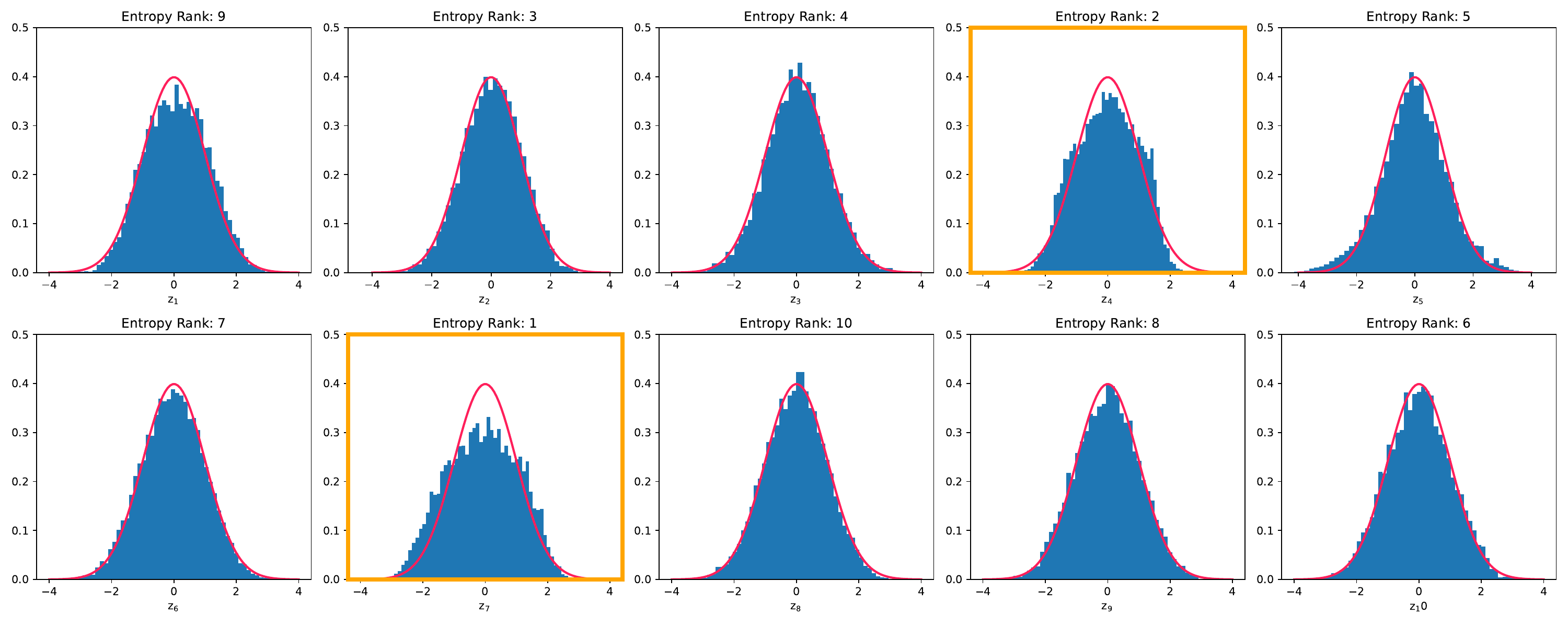}%
        \label{fig:histec}}
    \caption{Comparison of latent variable distributions between VAE and EC-VAE.
    The histograms are plotted in density scale, with the red line representing the PDF of a normal distribution.
    The entropy order of each latent variable is displayed at the top of each histogram, and the top two variables with the lowest entropy are highlighted in the yellow box.}
    \label{fig:comhist}
\end{figure}

Figure~\ref{fig:comhist} illustrates the distribution of latent variables. 
Without the entropy constraint, all latent variables in VAE follow a normal distribution. 
In contrast, EC-VAE encourages higher entropy in latent variables (Fig.~\ref{fig:histec}), stabilizing those selected for target embedding and resulting in clearly distinguishable low-entropy variables.

\begin{figure}[htb]
    \centering
    \subfloat[Correlation Heatmap: Latent Variables and Data Features in VAE]{%
        \includegraphics[width=0.45\linewidth]{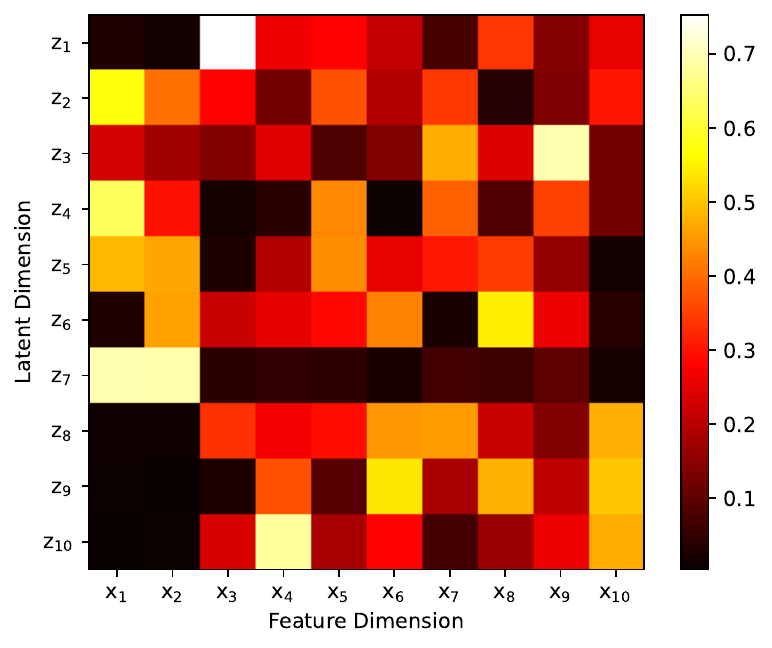}%
        \label{fig:corrvae}}
    \hfil
    \subfloat[Correlation Heatmap: Latent Variables and Data Features in EC-VAE]{%
        \includegraphics[width=0.45\linewidth]{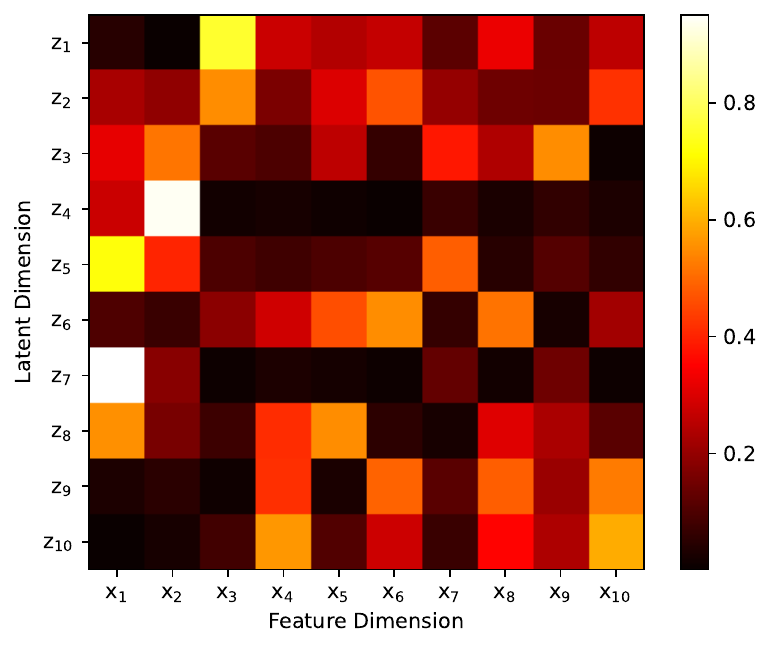}%
        \label{fig:correc}}
    \caption{Comparison of data feature encoding between VAE and EC-VAE using correlation heatmaps.
    Higher correlation values indicate that the latent variables more effectively encode the corresponding data features.}
    \label{fig:comcorr}
\end{figure}

Figure~\ref{fig:comcorr} further demonstrates that EC-VAE achieves better disentanglement of data features compared to VAE. 
However, selected latent variables in VAE also capture some target data features. 
This is reflected in Table~\ref{tbl:s1losses}, where the VAE's regressor exhibits slightly better performance.

\begin{table}[htb]
\caption{Average Losses of VAE and EC-VAE at Stage 1}
\label{tbl:s1losses}
\centering
\begin{tabular}{lcccc}
\hline
\textbf{Model}  & \textbf{Reconstruction} & \textbf{KL} & \textbf{Entropy} & \textbf{Regressor} \\ \hline
VAE         & \textbf{2.473e-2  }       & 1.798e-3 & 1.281e-4  & \textbf{6.505e-2}    \\
EC-VAE      & 2.500e-2         & \textbf{1.783e-3} & \textbf{1.263e-4}  & 7.250e-2   \\
\hline
\end{tabular}
\end{table}

\subsubsection{UT Module and Refined Latent Representation}
After processing by the UT module, the reconstruction and regressor losses in Stage 2 reach 1.072e-2 and 8.130e-3, reductions of 57\% and 89\% relative to Stage 1 in Table~\ref{tbl:s1losses}.
\begin{table}[htb]
\caption{Statistics of the Regressors in Stage 1 and Stage 2}
\label{tbl:regstat}
\centering
\begin{tabular}{lcccccc}
\hline
        & \multicolumn{2}{c}{$\hat{y}$} & \multicolumn{2}{c}{$\hat{c}_E$} & \multicolumn{2}{c}{$\hat{c}_I$} \\
\cline{2-7}
Stage   & \textbf{Mean}  & \textbf{Std.}  & \textbf{Mean}  & \textbf{Std.}  & \textbf{Mean}  & \textbf{Std.}  \\ \hline
Stage 1 & 4.771         & 2.168        & -0.1856       & 2.071       & 0.2148        & 2.0831       \\
Stage 2 & 1.542         & 1.0204       & -0.08448      & 0.9652      & 0.1367        & 0.9634      \\
\hline
\end{tabular}
\end{table}
Table~\ref{tbl:regstat} compares the regression statistics between Stage 1 (without UT module) and Stage 2 (with UT module).
Because the target features were normalized during training, a well-calibrated regressor should reproduce predictions with mean near $0$ and standard deviation near $1$; Stage 2 is closer to both than Stage 1 on all three targets.
We note that this measures distributional calibration rather than predictive accuracy.
\begin{figure*}[htb]
    \centering
    \subfloat[Selected Latent Variables vs. $\hat{y}$ in Stage 1]{%
        \includegraphics[width=0.3\linewidth]{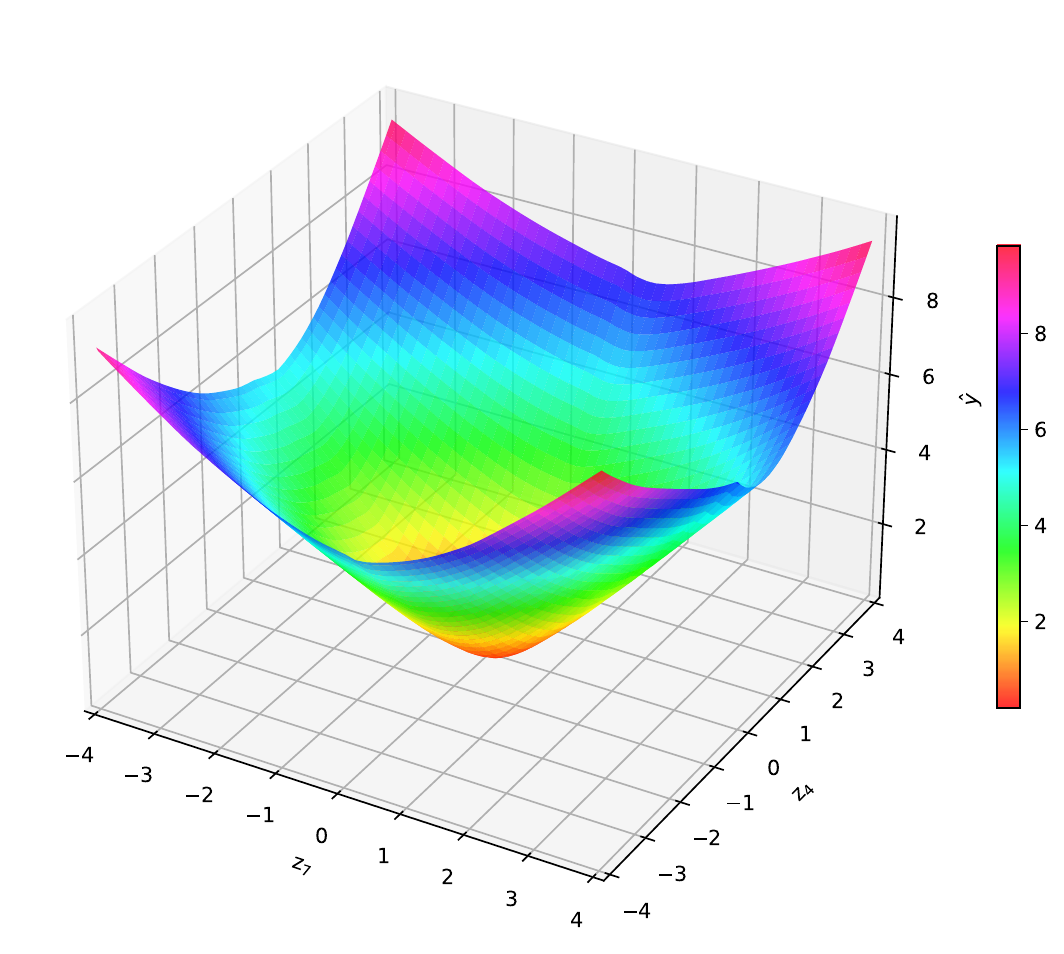}
        \label{fig:s1_3d_y}}
    \hfil
    \subfloat[Selected Latent Variables vs. $\hat{c}_E$ in Stage 1]{%
        \includegraphics[width=0.3\linewidth]{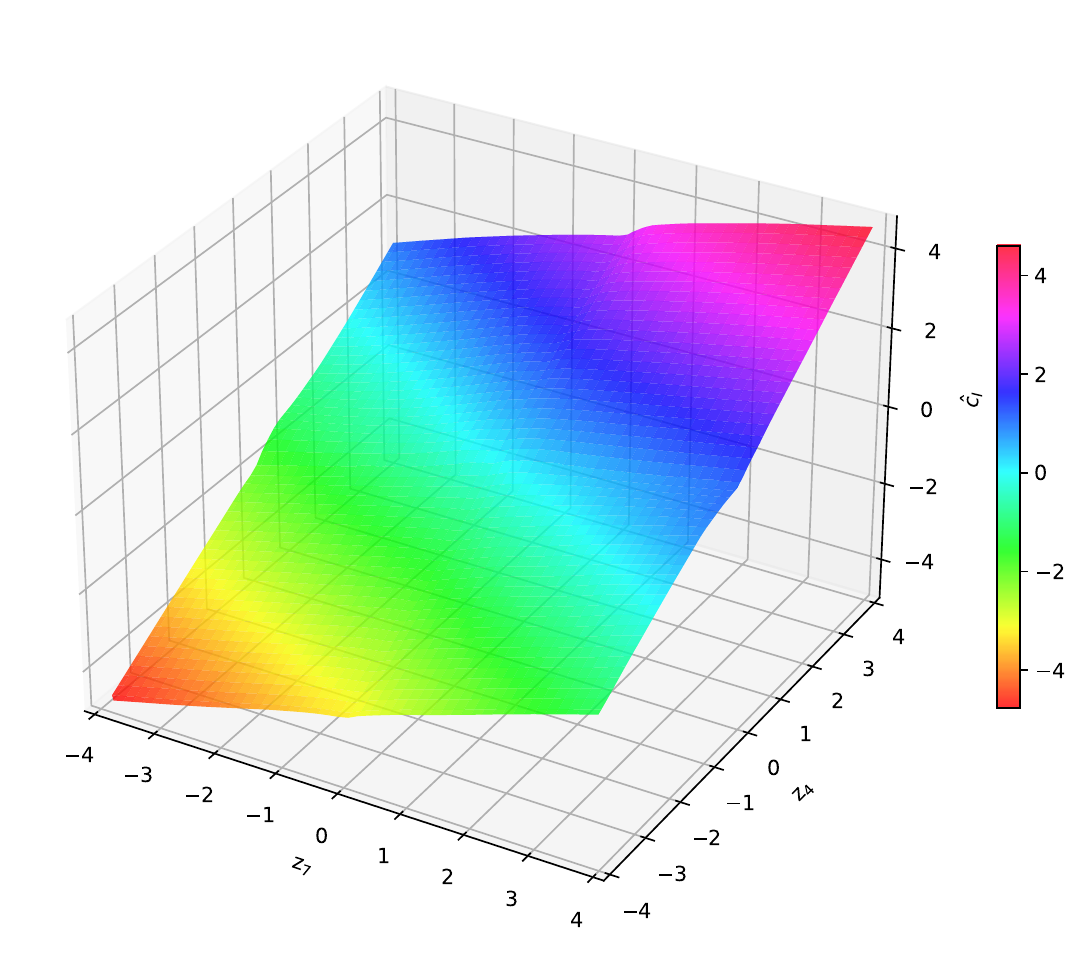}
        \label{fig:s1_3d_cE}}
    \hfil
    \subfloat[Selected Latent Variables vs. $\hat{c}_I$ in Stage 1]{%
        \includegraphics[width=0.3\linewidth]{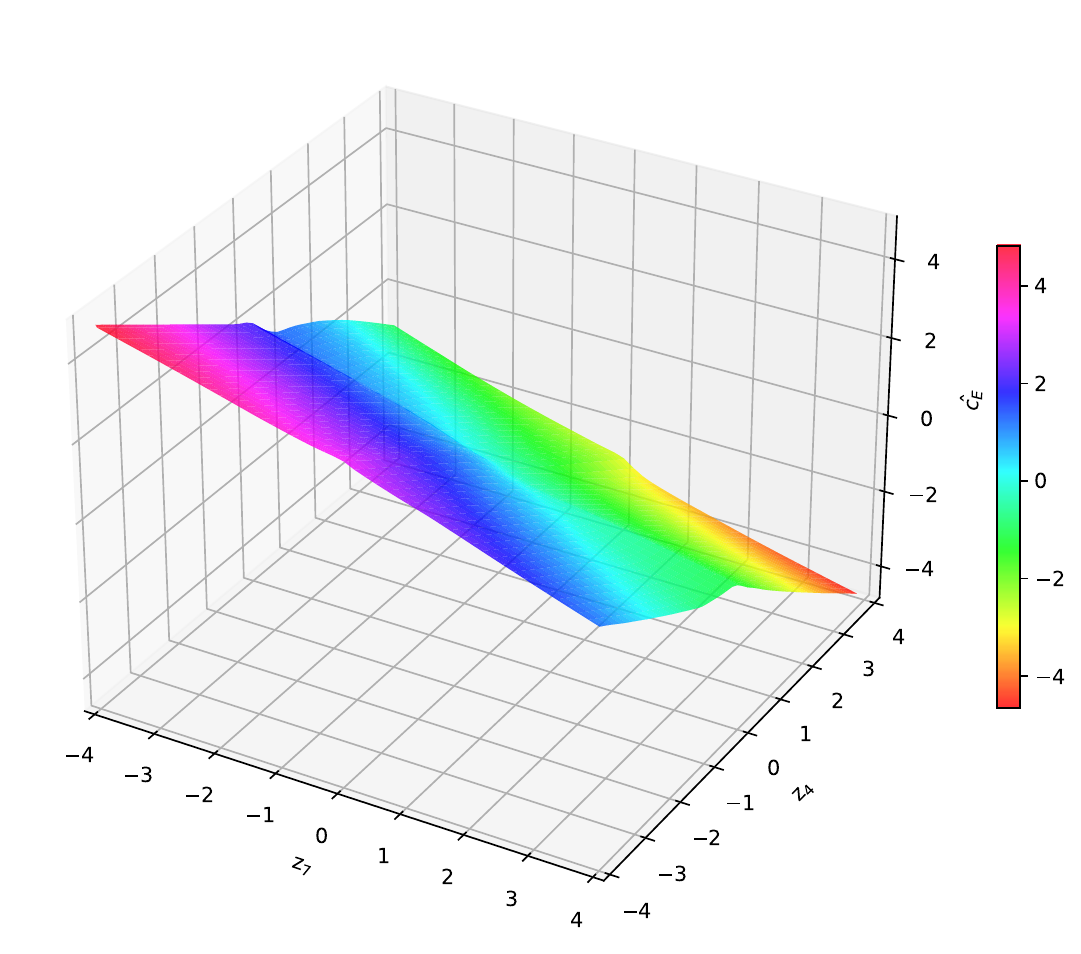}
        \label{fig:s1_3d_cI}}
    \vfil
    \subfloat[Selected Latent Variables vs. $\hat{y}$ in Stage 2]{%
        \includegraphics[width=0.3\linewidth]{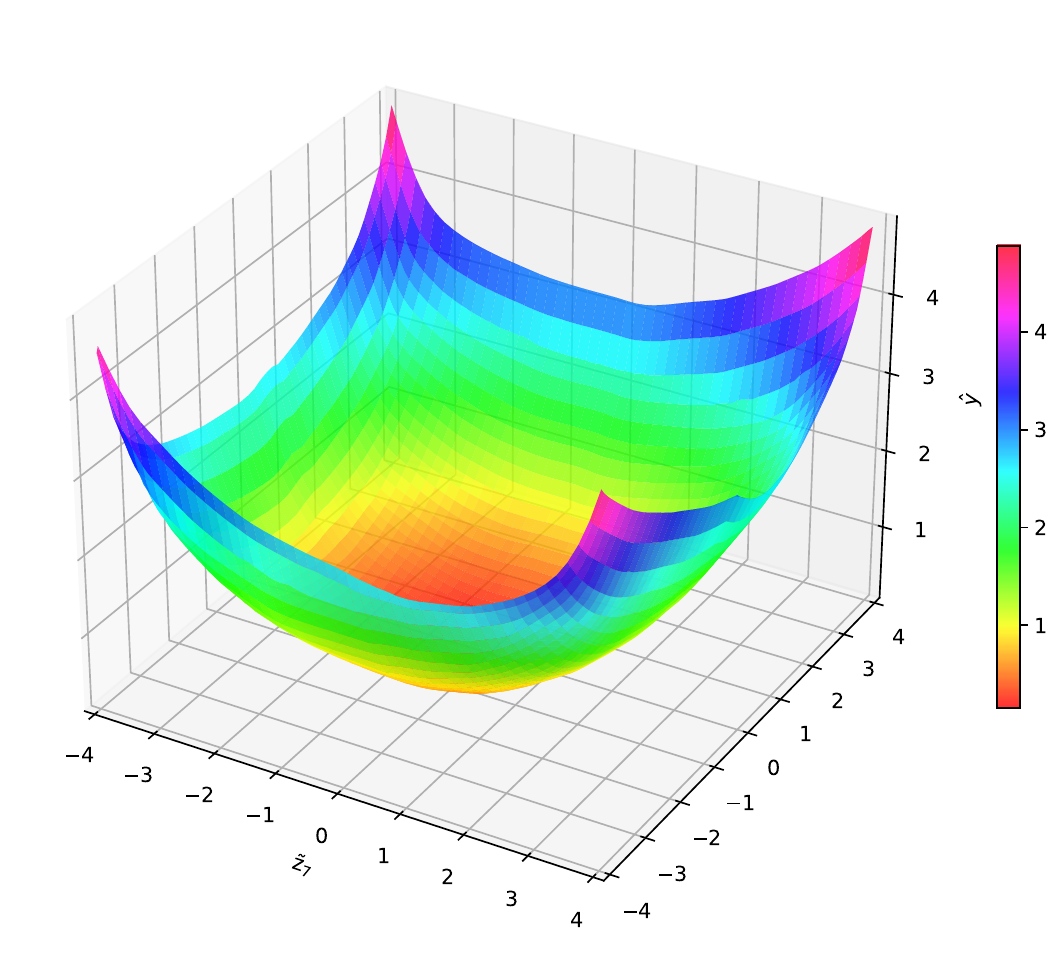}
        \label{fig:s2_3d_y}}
    \hfil
    \subfloat[Selected Latent Variables vs. $\hat{c}_E$ in Stage 2]{%
        \includegraphics[width=0.3\linewidth]{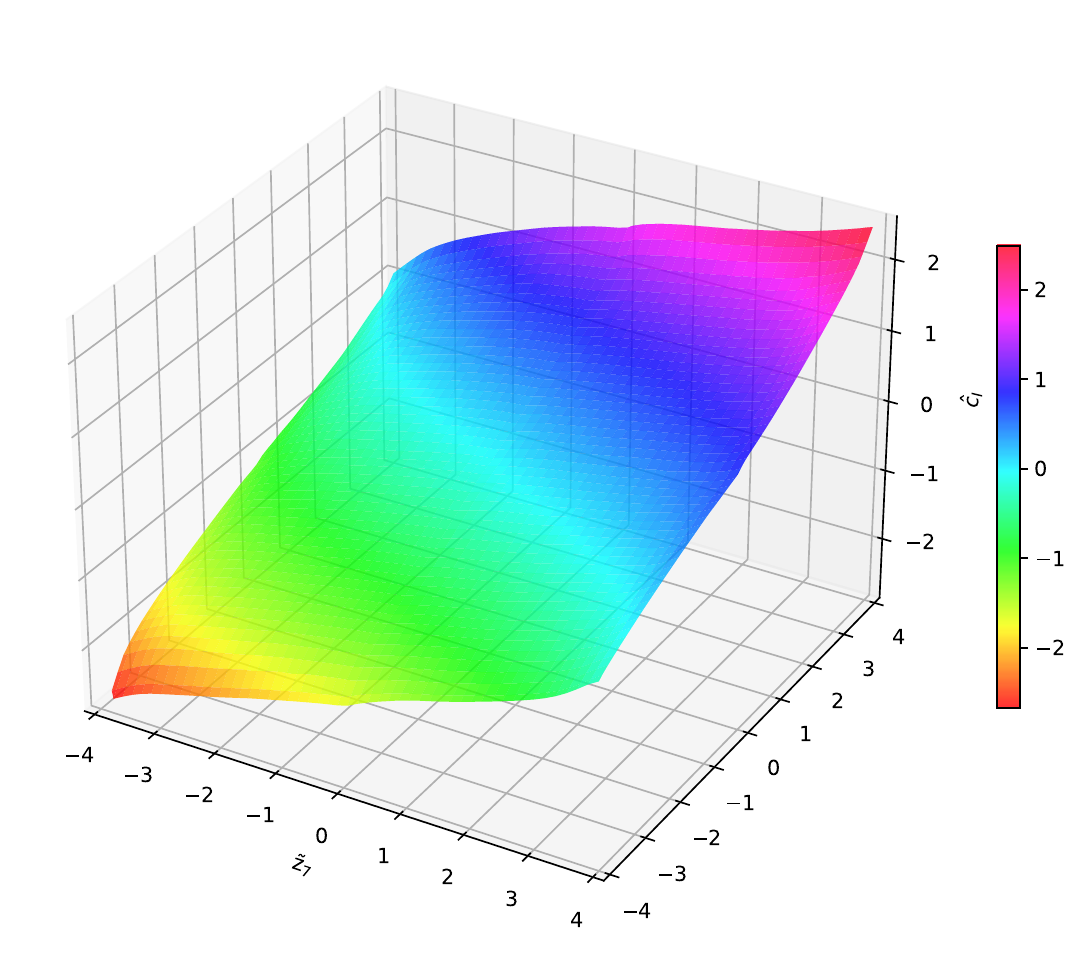}
        \label{fig:s2_3d_cE}}
    \hfil
    \subfloat[Selected Latent Variables vs. $\hat{c}_I$ in Stage 2]{%
        \includegraphics[width=0.3\linewidth]{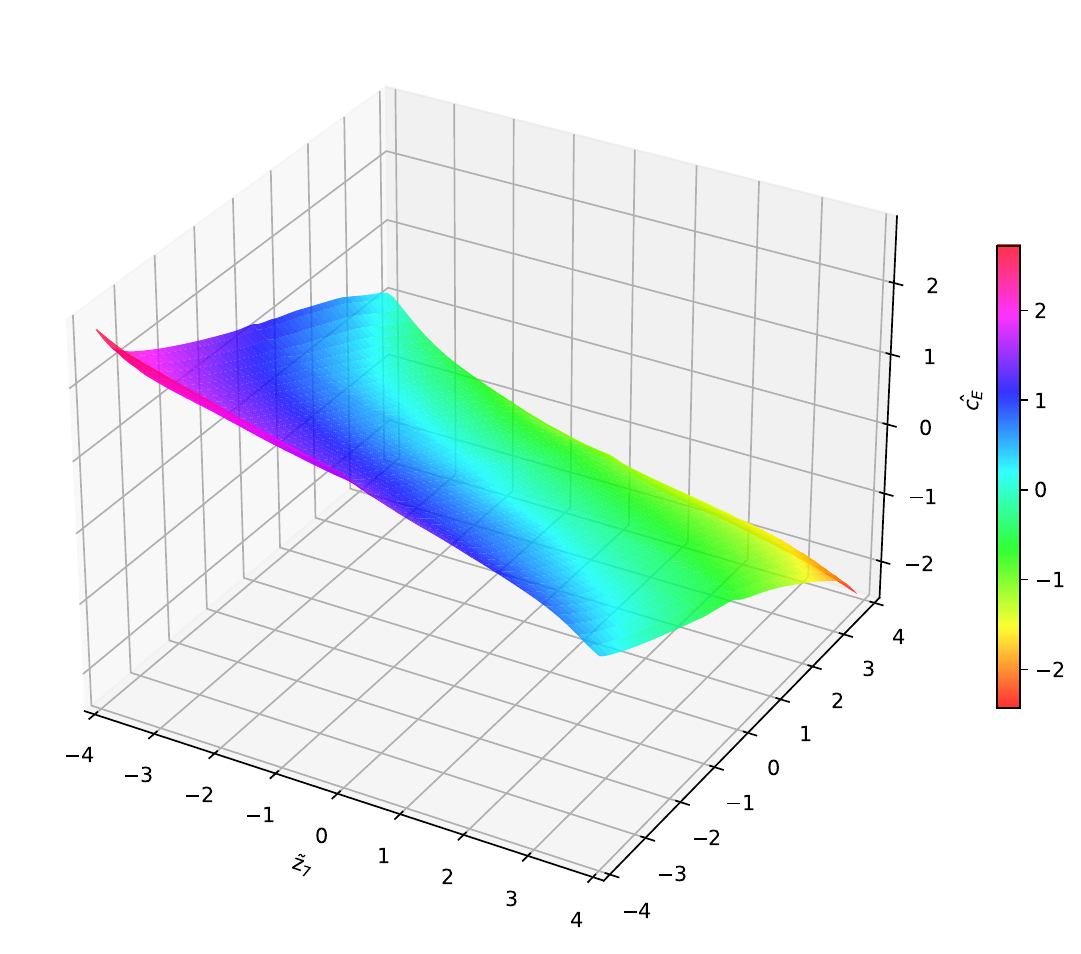}
        \label{fig:s2_3d_cI}}

    \caption{Comparison of learned manifolds between Stage 1 and Stage 2.
    The first row presents the manifolds of the regressor in Stage 1, while the second row presents the manifolds in Stage 2.
    The x-axis and y-axis represent the selected latent variables, and the z-axis corresponds to the predicted values $\hat{y}$, $\hat{c}_E$, and $\hat{c}_I$.
    The refined latent representation in Stage 2 results in smoother and more structured manifolds, demonstrating improved feature disentanglement.}
    \label{fig:manifolds}
\end{figure*}
Figure~\ref{fig:manifolds} compares the learned manifolds between Stage 1 and Stage 2. 
In Stage 1 (Figs.~\ref{fig:s1_3d_cE} and \ref{fig:s1_3d_cI}), the manifolds exhibit wrinkles (non-smooth, high-curvature regions) caused by the GM assumptions in the VAE architecture. 
Our UT module resolves these wrinkles, resulting in smoother manifolds in Stage 2 (Figs.~\ref{fig:s2_3d_cE} and \ref{fig:s2_3d_cI}) while retaining the convexity of the original optimization problem.

\subsubsection{Penalty Method vs. CPFM}

We compare our CPFM approach with the penalty method, which uses fixed Lagrangian multipliers. 
The loss function for the penalty method is defined as:
\begin{equation}
    L_p(\zeta) = y(\zeta) - c_E(\zeta) + c_I(\zeta).
\end{equation}

\begin{figure}[htb]
    \centering
    \subfloat[Convergence of the Penalty Method]{%
        \includegraphics[width=0.45\linewidth]{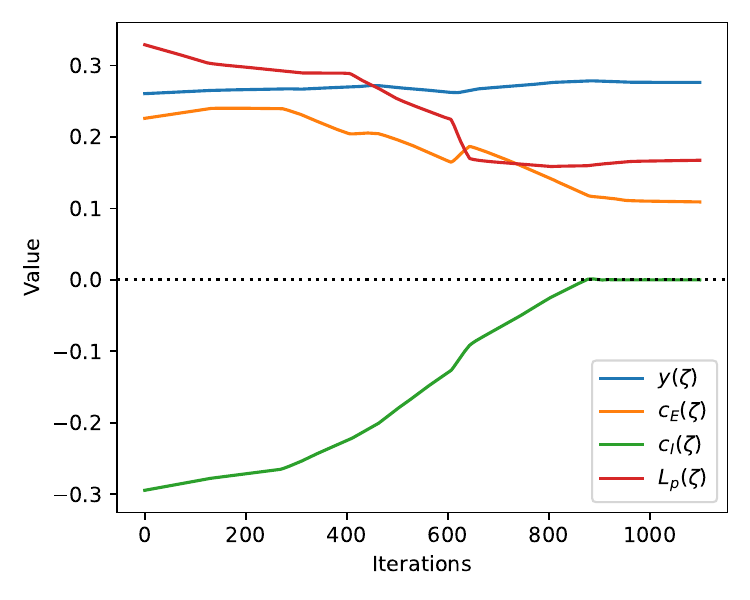}
        \label{fig:cvg_panelity}}
    \hfil
    \subfloat[Convergence of CPFM]{%
        \includegraphics[width=0.45\linewidth]{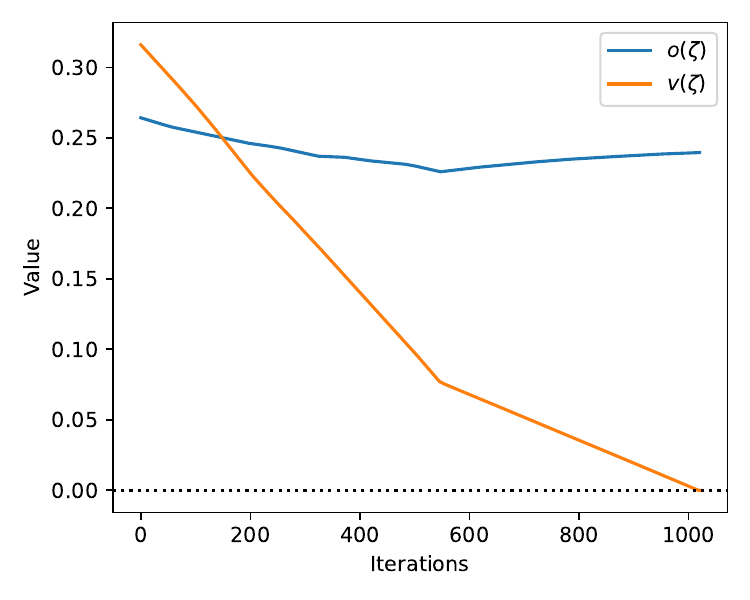}
        \label{fig:cvg_CPFM}}

    \caption{Comparison of convergence behaviors during the optimization process over 1,000 iterations.}
    \label{fig:cvg}
\end{figure}

Figure~\ref{fig:cvg} illustrates the convergence behavior of both methods over 1,000 iterations. 
While both approaches achieve convergence, the equality constraint $c_E$ in Figure~\ref{fig:cvg_panelity} remains unsatisfied, exhibiting a persistent offset above zero. 

However, the violation function $v(\zeta)$ in Figure~\ref{fig:cvg_CPFM} ultimately converges to zero, indicating that all constraints are fully satisfied. 
This result demonstrates that CPFM effectively enforces hard constraints in the optimization process.

\subsection{Drug Discovery Problem Setup}
We formalize the drug discovery problem using the ZINC250k dataset, a curated subset of the broader ZINC database~\cite{irwinZINC20AFreeUltralargeScale2020} containing 250,000 drug-like molecules often employed in computational chemistry and machine learning research. 
Each molecule is represented as a SMILES string~\cite{weiningerSMILESChemicalLanguage1988}, which is inherently discrete. 
Therefore, the resulting optimization problem is \emph{mixed-integer} in nature, as it involves selecting discrete tokens and potentially variable-length sequences to satisfy both structural and pharmacological requirements.

Let $\rvx$ denote a SMILES string of length up to 128, constructed from a vocabulary of 37 discrete tokens. 
We aim to find structurally and pharmacologically favorable compounds by minimizing the molecular weight, $\mathrm{M_{wt}}(\rvx)$, subject to two key constraints: 
the logP value of the molecule, $\mathrm{logP}(x)$, must lie between 1 and 3, 
and the reactivity, $\mathrm{reactivity}(x)$, must be zero. 
Formally, we write:
\begin{equation}\label{eqn:zinc-problem}
\begin{aligned}
\min_{\rvx \in \{0,1\}^{128 \times 37}} \quad & \mathrm{M_{wt}}(\rvx), \\
\text{subject to} \quad & 1 \,\leq\, \mathrm{logP}(\rvx) \,\leq\, 3, \\
& \mathrm{reactivity}(\rvx) \;=\; 0.
\end{aligned}
\end{equation}

These constraints embody common drug-likeness criteria. 
Constraining logP between 1 and 3 aims to achieve balanced lipophilicity, which generally fosters effective absorption and distribution. 
Setting reactivity to zero ensures that the designed compounds are chemically stable and less prone to unwanted reactions, thereby reducing the likelihood of toxicity. 
Finally, minimizing the molecular weight helps promote solubility and metabolic stability, both critical properties for successful drug candidates.

To encode each molecule into a continuous latent space, we employ a 4-layer Transformer~\cite{vaswaniAttentionAllYou2017} encoder with 8 attention heads and an embedding size of 64.
This encoder outputs a latent vector of dimension 256. 
A matching 4-layer Transformer decoder is used for reconstructing SMILES strings from the latent space. 
The multipliers for the EC-VAE are $\beta = 0.3$ and $\gamma = 1$. 

We further embed a three-layer MLP regressor (hidden dimension 3,000) to map the 256-dimensional latent representation to the normalized target features:
$$
\bigl[\mathrm{M_{wt}},\;\mathrm{logP},\;\mathrm{reactivity}\bigr]^\mathsf{T}.
$$
We use the same tolerance settings in our optimization algorithm (Algorithm~\ref{alg:CPFM}) as in the synthetic problem. 
The problem contains one equality constraint and a two-sided bound on logP, so following \eqref{eqn:violation} we write the violation function of CPFM as
\begin{align}\label{eqn:violation-drug}
v(\zeta) 
&= \bigl\|\widehat{\mathrm{reactivity}}(\zeta)\bigr\|_2^{2}
    + \max\!\Bigl(0,\;1 - \widehat{\mathrm{logP}}(\zeta)\Bigr)^{2} \nonumber\\
&\quad + \max\!\Bigl(0,\;\widehat{\mathrm{logP}}(\zeta) - 3\Bigr)^{2}.
\end{align}
Each rectifier is active only outside its own bound, so $v(\zeta) = 0$ exactly on the feasible set: at $\widehat{\mathrm{logP}} = 2$ with zero predicted reactivity, $v = 0$, whereas $\widehat{\mathrm{logP}} = 0.5$ and $\widehat{\mathrm{logP}} = 3.5$ both give $v = 0.25$.

By organizing the data into a surrogate model in latent space, this approach enables efficient exploration of the drug-like chemical space while enforcing the mixed-integer constraints imposed by the discrete SMILES representation.

\subsection{Results of the Drug Discovery Problem}
We generated 10 candidate solutions for the drug discovery task. 
Because the feature selector retained 252 out of the original 256 latent dimensions, only a small fraction of latent space remained for sampling, thereby limiting the number of feasible solutions.
Table~\ref{tbl:drugcom} presents the averaged results of our MCOF compared to several established molecular optimization methods. 
For each generated molecule, we compute the molecular weight (MWt), the octanol--water partition coefficient (logP), and the quantitative estimate of drug-likeness (QED). 
Reactivity (Reac.) is reported from the Stage~2 MLP regressor rather than a direct calculation, as a suitable external measure is unavailable. 
The violation rate (Vio.) indicates how often the generated molecules fail the imposed constraints, 
and New reflects the percentage of molecules not present in the training set.
VAE+GA~\cite{gomez-bombarelliAutomaticChemicalDesign2018} was among the earliest to integrate a genetic algorithm (GA) with VAE-based latent representations in drug discovery.
VAE+BO~\cite{zhuBayesianDeepConvolutional2018} subsequently employed Bayesian Optimization (BO) in the latent space. 
HGNN~\cite{jinHierarchicalGraphtoGraphTranslation2019} replaced MLP components with graph neural networks (GNNs) to capture molecular structures more effectively. 
CG-VAE~\cite{liuConstrainedGraphVariational2018} imposed distribution constraints to enhance the validity of generated molecules, 
while MolDQN~\cite{zhouOptimizationMoleculesDeep2019a} applied reinforcement learning for constrained molecular optimization.

\begin{table}[htb]
\centering
\caption{Performance comparison among various methods for molecular optimization.}
\label{tbl:drugcom}
\begin{tabular}{lcccccc}
\hline
\textbf{Method} & \textbf{MWt} & \textbf{logP} & \textbf{Reac.} & \textbf{QED} & \textbf{Vio.} & \textbf{New} \\
\hline
MCOF (ours)                             & 295.57 & 2.98  & -0.013   & 0.58  & 0\%       & 100\%   \\
VAE+GA \cite{gomez-bombarelliAutomaticChemicalDesign2018} & --     & 1.46  & --  & 0.75  & --        & 93\%    \\
VAE+BO~\cite{zhuBayesianDeepConvolutional2018}                                  & --     & --    & --  & --    & 8\%       & 2.25\%  \\
HGNN \cite{jinHierarchicalGraphtoGraphTranslation2019}     & --     & 2.49  & --  & 0.47  & 26.4\%    & --      \\
CG-VAE \cite{liuConstrainedGraphVariational2018}           & --     & --    & --  & --    & --        & 99.82\% \\
MolDQN \cite{zhouOptimizationMoleculesDeep2019a}           & --     & 11.84 & --  & 0.859 & 0\%       & --      \\
\hline
\end{tabular}
\end{table}

Our MCOF achieves zero constraint violations (\textbf{Vio.} = 0\%), showing that it strictly enforces logP and reactivity bounds. 
In addition, all generated molecules are unique (\textbf{New} = 100\%), indicating high diversity in the discovered chemical space. 
While approaches like VAE+GA, VAE+BO, and CG-VAE also focus on exploiting latent representations, they either report nonzero violation rates or generate fewer novel structures.
Methods such as HGNN highlight the utility of graph architectures, whereas MolDQN relies on reinforcement learning to navigate discrete chemical spaces. 
By contrast, MCOF's multi-stage pipeline effectively balances constraint enforcement, molecular novelty, and complexity, making it a robust solution for data-driven drug discovery problems.

We selected the top three molecules out of the 100 generated samples, listed in Table~\ref{tbl:example_molecules} with their respective data. 
Note that lower (or more negative) reactivity values indicate lower chemical reactivity, while higher QED values are generally more desirable in drug design.
\begin{table}[htb]
\centering
\caption{Top three molecules selected from the generated set, alongside their key properties.}
\label{tbl:example_molecules}
\begin{tabular}{c l c c c c}
\hline
\textbf{\#} & \textbf{Molecule} & \textbf{MWt} & \textbf{logP} & \textbf{Reactive} & \textbf{QED} \\
\hline
1 & 
\includegraphics[width=0.3\columnwidth]{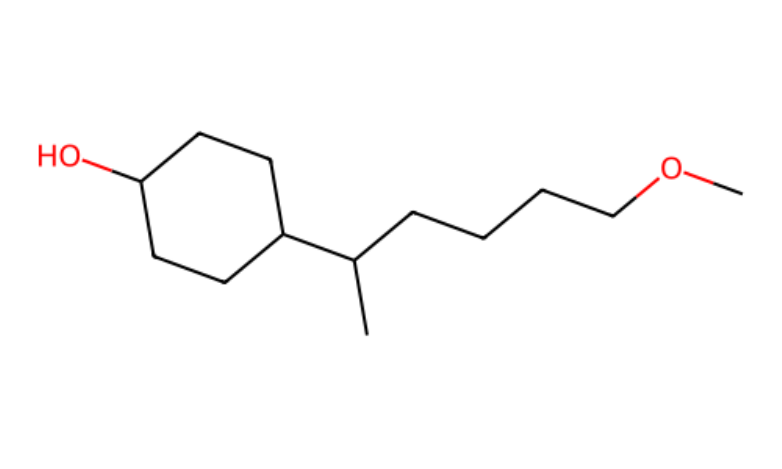} &
214.35 & 2.99 & -0.024 & 0.69 \\
2 & 
\includegraphics[width=0.3\columnwidth]{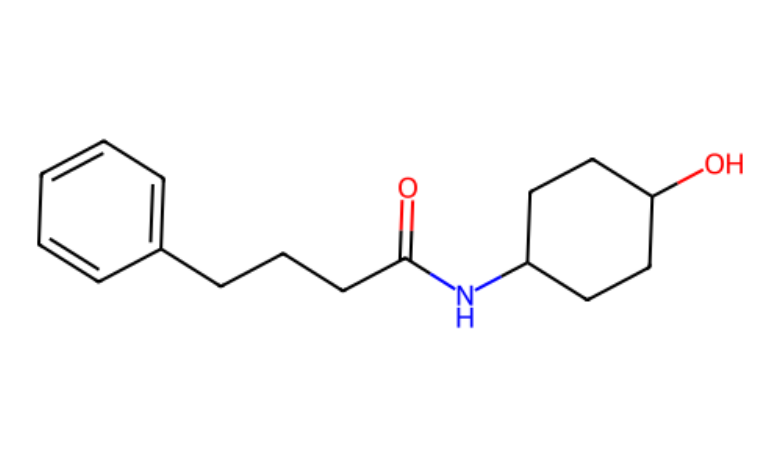} &
261.36 & 2.43 & -0.011 & 0.85 \\
3 & 
\includegraphics[width=0.3\columnwidth]{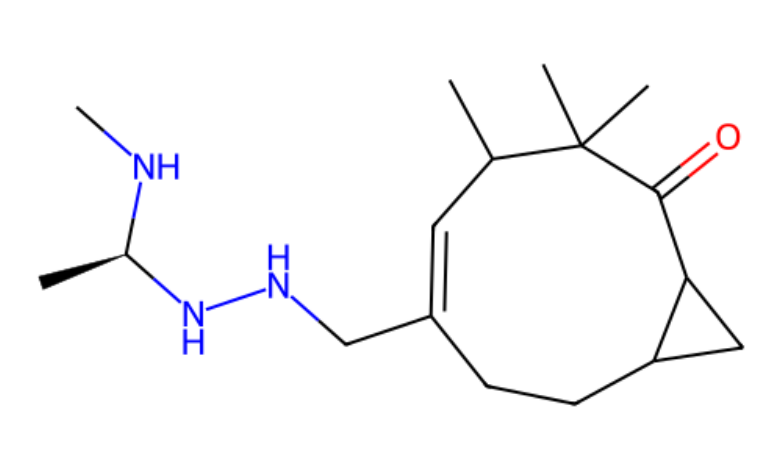} &
293.45 & 2.23 & -0.009 & 0.41 \\
\hline
\end{tabular}
\end{table}

In summary, these two experiments demonstrate our MCOF consistently produces hard-constrained valid solutions.
And our MCOF is not sensitive to numerical optimization and mixed-integer optimization for data-driven problems.

\section{Conclusion}\label{sec:conclusion}

This paper presented MCOF that leverages VAEs for data-driven black-box optimization.
Our framework addresses three major challenges often encountered in VAE-based optimization: effective sampling in latent space, identification of active decision variables, and imposition of hard constraints.

First, we integrated an EC-VAE with a feature selector to embed objective and constraint information into lower-entropy latent dimensions.
This stage helps differentiate and retain only the most salient factors of variation within the data. 
Second, a UT module was introduced to mitigate distributional mismatch issues by converting latent variables to a uniform distribution, thereby improving sampling consistency and decoder performance. 
Third, we developed a CPFM to systematically enforce hard constraints while concurrently minimizing the objective function, avoiding the potential instability of alternative approaches such as Lagrangian multipliers. 
Finally, we proposed a latent solution completion step to handle any residual, unselected latent variables.

Comprehensive experiments demonstrated the effectiveness of MCOF on both a synthetic problem and a realistic drug discovery task. 
In the synthetic case, our method consistently generated feasible solutions without constraint violations. 
In the drug discovery application, we showed that MCOF could discover highly diverse and logP-compliant molecules with low reactivity---outperforming or complementing existing VAE-based and reinforcement-learning-based approaches.

Moving forward, several enhancements can further strengthen the applicability and efficiency of MCOF. 
First, refining the feature selector's time complexity with advanced algorithms such as bucket sorting or binary searching can reduce computational overhead. 
Second, devising more efficient techniques to estimate latent variable entropy would allow EC-VAE to accommodate even higher-dimensional datasets or data with mixed variable types. 
Lastly, exploring broader problem classes---such as multi-objective optimization, combinatorial constraints, or domains beyond drug discovery---would help validate the generality and robustness of our approach. 
We believe that these directions will empower MCOF to serve a wider range of data-driven optimization tasks across engineering and scientific applications.

\ifCLASSOPTIONcaptionsoff
  \newpage
\fi

\appendices

\bibliographystyle{IEEEtran}

\bibliography{mybibfile}

\end{document}